\documentclass[10pt,twocolumn,letterpaper]{article}
\usepackage[accsupp]{axessibility}

\usepackage{iccv}              

\definecolor{iccvblue}{rgb}{0.21,0.49,0.74}
\definecolor{hotmagenta}{RGB}{255, 51, 153}
\usepackage[pagebackref,breaklinks,colorlinks]{hyperref}

\hypersetup{
  urlcolor=hotmagenta,
  linkcolor=iccvblue,
  citecolor=iccvblue
}

\usepackage{multirow}
\usepackage{float}
\usepackage[most]{tcolorbox}

\usepackage{algorithm}
\usepackage{algpseudocode}
\usepackage{amsmath}
\usepackage{makecell}
\usepackage{colortbl}
\definecolor{LightOrange}{rgb}{1,0.85,0.8}
\definecolor{blond}{rgb}{0.98, 0.94, 0.75}
\definecolor{blizzardblue}{rgb}{0.67, 0.9, 0.93}
\definecolor{LightGreen}{rgb}{0.93,0.98,0.96}
\definecolor{babypink}{rgb}{0.96, 0.76, 0.76}
\definecolor{classicrose}{rgb}{0.98, 0.8, 0.91}
\definecolor{textboxblue}{RGB}{22, 98, 132}
\definecolor{textboxgrey}{RGB}{252, 252, 252}
\definecolor{commentgreen}{RGB}{25, 107, 36}
\definecolor{lightblue}{rgb}{0.18,0.45,0.71} 

\newcommand{\blank}[1]{\makebox[#1][l]{}}

\def\paperID{9168} 
\def\confName{ICCV}
\def\confYear{2025}

\title{T2I-Copilot: A Training-Free Multi-Agent Text-to-Image System for\\ Enhanced Prompt Interpretation and Interactive Generation}

\author{
    Chieh-Yun Chen, 
    Min Shi, 
    Gong Zhang, 
    Humphrey Shi \\ 
{\small SHI Labs @ Georgia Tech} \\
\centering{{\small \textbf{\texttt{\href{https://github.com/SHI-Labs/T2I-Copilot}{github.com/SHI-Labs/T2I-Copilot}}}}}
}

\begin{document}

\twocolumn[{%
    \maketitle
        \vspace{-1cm}
        \begin{figure}[H]
        \hsize=\textwidth
        \includegraphics[width=0.95\textwidth]{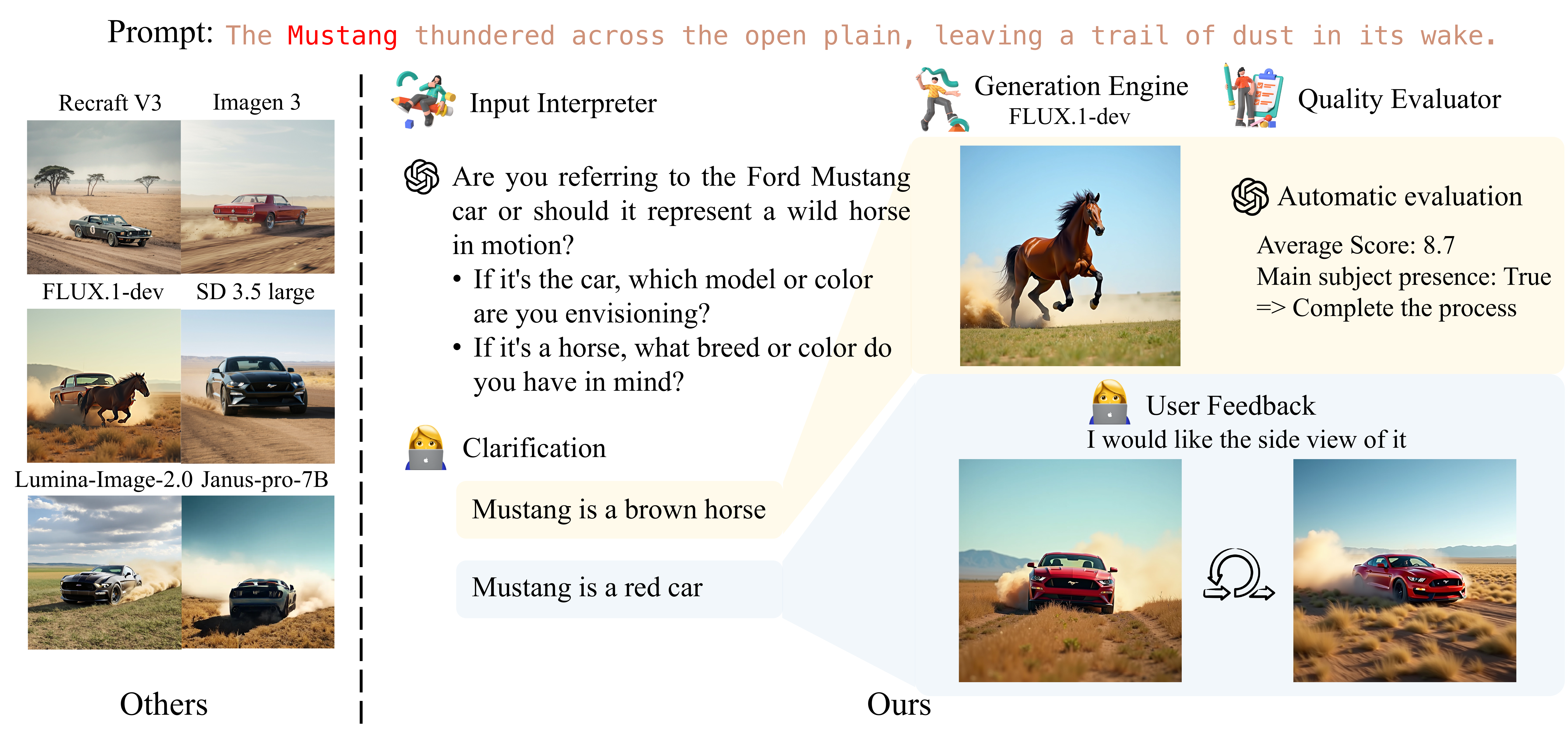}
        \centering
        \vspace{-.3cm}
        \caption{\textbf{T2I-Copilot: An interactive agentic Text-to-Image generation system.}
        Current generative models struggle to interpret complex or ambiguous user prompts, often failing to produce images that perfectly align with user intent. We propose a multi-agent system that refines input prompts, resolves ambiguities, and iteratively evaluates results, providing feedback to guide regeneration when needed. Users can supplement information interactively, or large language models can do so autonomously. Our approach enhances both aesthetics and text-image alignment without requiring additional training or intricate prompt engineering.
        }
    \label{fig:fig-teaser-Mustang}
    \end{figure}
}]


\begin{abstract}
Text-to-Image (T2I) generative models have revolutionized content creation but remain highly sensitive to prompt phrasing, often requiring users to repeatedly refine prompts multiple times without clear feedback. 
While techniques such as automatic prompt engineering, controlled text embeddings, denoising, and multi-turn generation mitigate these issues, they offer limited controllability, or often necessitate additional training, restricting the generalization abilities.
Thus, we introduce T2I-Copilot, a training-free multi-agent system that leverages collaboration between (Multimodal) Large Language Models to automate prompt phrasing, model selection, and iterative refinement. This approach significantly simplifies prompt engineering while enhancing generation quality and text-image alignment compared to direct generation.
Specifically, T2I-Copilot consists of three agents: (1) Input Interpreter, which parses the input prompt, resolves ambiguities, and generates a standardized report; (2) Generation Engine, which selects the appropriate model from different types of T2I models and organizes visual and textual prompts to initiate generation; and (3) Quality Evaluator, which assesses aesthetic quality and text-image alignment, providing scores and feedback for potential regeneration.
T2I-Copilot can operate fully autonomously while also supporting human-in-the-loop intervention for fine-grained control.
On GenAI-Bench, using open-source generation models, T2I-Copilot achieves a VQA score comparable to commercial models RecraftV3 and Imagen 3, surpasses FLUX1.1-pro by 6.17\% at only 16.59\% of its cost, and outperforms FLUX.1-dev and SD 3.5 Large by 9.11\% and 6.36\%. Code will be released at: \href{https://github.com/SHI-Labs/T2I-Copilot}{github.com/SHI-Labs/T2I-Copilot}.

\end{abstract}    
\section{Introduction}
\label{sec:intro}

Although Text-to-Image (T2I) generative models~\cite{imagen3_24, recraftv3_24, flux24, midjourneyv6.1_24, dalle3_24, sdxl_ICLR24, luminaImage_25, janusPro7B_25, finestyle_NeurIPS24} have made significant strides in generating realistic images across various styles, they remain highly sensitive to prompt phrasing.  If a prompt is ambiguous or casually written, the model may fail to generate images that fully align with the user's intent. Inexperienced users may repeatedly refine prompts without achieving the desired results. This challenge arises because, unlike Large Language Models (LLMs), which communicate directly with users in natural language to explicitly address their needs, T2I models rely on text prompts enforced through a text encoder during generation. These models do not offer reasoning or analysis like LLMs and do not provide direct feedback on their internal understanding or knowledge gaps when generation fails. \textbf{This lack of interpretability complicates error analysis and refinement.} For instance, as shown in Fig.~\ref{fig:fig-teaser-Mustang}, when given the prompt ``The Mustang thundered across the open plain, leaving a trail of dust in its wake,'' most T2I models~\cite{imagen3_24, recraftv3_24, SD35_24, luminaImage_25, janusPro7B_25} predominantly generate an image of a car, whereas FLUX.1-dev~\cite{flux24} generates both a horse and a car. This ambiguity arises because models either fail to recognize that \textit{Mustang} can refer to both a car and a horse or lacks sufficient contextual cues to align with user intent. As a result, users must spend considerable computational resources and time refining prompts, sampling multiple times, or even fine-tuning models to achieve better results.

Existing approaches have attempted to address these limitations through various strategies, including enhanced prompt engineering~\cite{Po_NeurIPS23, metaPO_TMLR24}, control mechanisms within text embeddings~\cite{CAT_NeurIPS24, tome_NeurIPS24} or the denoising process~\cite{A&E_siggraph23, syngen_nips23, goldennoise_24arxiv}, leveraging LLMs for regional coordination~\cite{LMD_TMLR24, RAG_arxiv24, omost_24}, and multi-turn self-enhancement with tools~\cite{SLD_CVPR24, Genartist_NeurIPS24,google_multiturn24}. However, two more key challenges persist: 
Firstly, \textbf{trade-offs between architectural modifications and generalization.} Improvements in attribute binding~\cite{A&E_siggraph23, syngen_nips23} or de-bias~\cite{CAT_NeurIPS24} enhance specific use cases but often reduce flexibility in handling diverse prompts. 
Secondly, \textbf{limited user controllability.} For instance, PASTA~\cite{google_multiturn24} introduces a multi-turn generation approach where users select their preferred image. However, it lacks fine-grained control, preventing users from specifying details like object attributes or style adjustments. It also does not address cases where none of the generated images match user intent, limiting interaction to selection rather than refinement. Similarly, GenArtist~\cite{Genartist_NeurIPS24} selects a model from its predefined eighteen tools based solely on the prompt, without analyzing user intent beforehand. While it uses LLMs to predict bounding boxes for prompt-specified objects, this interpretation applies only to specific tools like LMD~\cite{LMD_TMLR24} and does not generalize across the system. Moreover, it lacks human-in-the-loop interaction, restricting user control over generation. 
Currently, there is no unified system that integrates comprehensive functionalities: i) enhance input interpretation before generation, ii) generate images using multiple tools without fine-tuning or architectural modifications, and iii) iterative self-improvement via multi-turn interactions.

To address above-mentioned three key challenges in T2I generation, we propose T2I-Copilot, a training-free multi-agent system that enhances controllability and interpretability by proactively analyzing user intent, selecting the optimal model, and iteratively refining results. The system consists of three sequential agents: i) \textbf{Input Interpreter Agent}: Analyzes user input by identifying key subjects, attributes (\eg, color, position), and image settings (\eg, background, style, lighting, camera parameters). It detects ambiguities which are clarified by MLLM creatively fill or user clarification when necessary and structures the analysis into a JSON-formatted report for precise generation. ii) \textbf{Generation Engine Agent}: Selects and executes the most suitable model based on user intent and model capabilities, enabling fine-grained control via Referring Expression Segmentation or an interactive user drawing canvas for targeted modifications. iii) \textbf{Quality Evaluator Agent}: Assesses the generated image based on aesthetic and text-image alignment criteria and provides improvement suggestions. If necessary, it incorporates user feedback and initiates iterative refinement. 
T2I-Copilot is a training-free framework, ensuring compatibility and scalability with the latest T2I models while integrating a human-in-the-loop approach for enhanced user control. Our contributions are as follows:
\begin{itemize}
    \item Proposing a training-free multi-agent system for T2I generation, where three specialized agents collaborate to improve model interpretability and generation efficiency.
    \item Bridging human intent and AI-driven creativity, enabling a more interpretable and interactive generative AI system.
    \item Achieving strong performance comparable to proprietary models like Recraft V3~\cite{recraftv3_24} and Imagen 3~\cite{imagen3_24}, while surpassing FLUX1.1-pro~\cite{flux24} by 6.17\% at only 16.59\%~\footnote{The cost comparison is detailed in Supplement D.} of its cost and outperforming the open-source FLUX.1-dev~\cite{flux24} and SD 3.5 Large~\cite{SD35_24} by 9.11\% and 6.36\%, respectively, with a VQAScore~\cite{VQAScore_ECCV24} on GenAI-Bench~\cite{GenAIBench_CVPRW24}.
\end{itemize}
\section{Related Works}
\label{sec:related_work}

\begin{figure*}[t]
  \centering
  \includegraphics[width=1.0\linewidth]{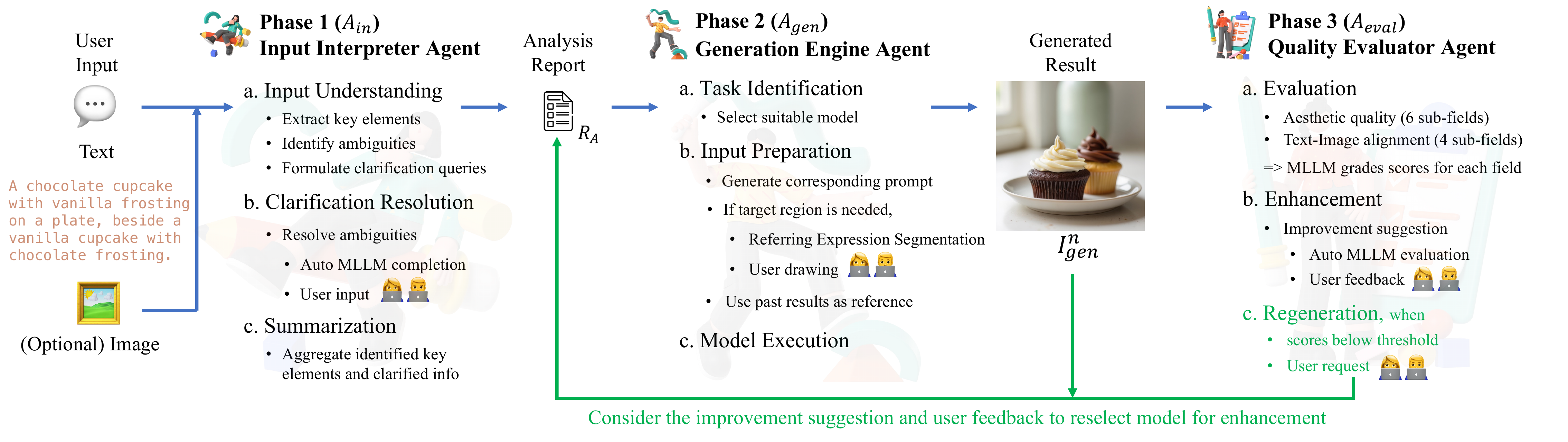}
  \vspace{-.5cm}
   \caption{\textbf{Pipeline of the proposed T2I-Copilot:} A multi-agent system for interactive text-to-image generation. The system consists of three sequential agents: (1) Input Interpreter processes user inputs, identifying ambiguities and either prompting the user for clarification or leveraging an MLLM for automatic refinement. (2) Generation Engine selects and executes the most suitable model based on the analysis report, user intent, and model capabilities. It enables fine-grained control via Referring Expression Segmentation or an interactive drawing canvas for targeted modifications. (3) Quality Evaluator assesses the generated image against aesthetic and alignment criteria, allowing user feedback to refine the output. If the image does not meet expectations, the system triggers regeneration to ensure improved results.}  
   \label{fig:pipeline}
\end{figure*}

\subsection{MLLM Agent}

Large Language Models (LLMs) have been increasingly leveraged as AI agents for complex tasks, including reasoning and decision-making~\cite{react_ICLR23}, tool utilization~\cite{huggingGPT_NeurIPS23}, and multi-agent collaboration~\cite{agentLab_25}. With the integration of vision capabilities, Multimodal Large Language Models (MLLMs) further extend these functionalities, making them valuable in T2I generation. Specifically, (M)LLMs have been employed in prompt engineering~\cite{metaPO_TMLR24, google_multiturn24}, self-correction and verification~\cite{SLD_CVPR24, Genartist_NeurIPS24}, prompt decomposition into object bounding boxes~\cite{LMD_TMLR24, RAG_arxiv24, omost_24, Genartist_NeurIPS24}, and model selection~\cite{visualchatgpt_23, Genartist_NeurIPS24, diffusionGPT_24arxiv}. 
Despite these advancements, existing approaches lack a unified framework that seamlessly integrates prompt interpretation, model selection, and iterative self-enhancement in T2I generation. 
To address this, we propose a proactive MLLM-driven multi-agent system that seamlessly unifies these processes, improving both controllability and effectiveness in image generation.

\subsection{Multi-turn Generation}

To better align T2I generation with user intent, several works have explored multi-turn approaches~\cite{SLD_CVPR24, Genartist_NeurIPS24, google_multiturn24}. 
SLD~\cite{SLD_CVPR24} employs an LLM to provide object coordinate modifications, allowing control over positioning and attributes.  
GenArtist~\cite{Genartist_NeurIPS24} leverages an MLLM for image verification and self-correction; however, in our reproduction of the publicly released code, we found that its self-corrections frequently diverge from the intended prompt, as shown in Fig.~\ref{fig:eval-and-regen-cp-Genartist}. In addition, it lacks support for user feedback. 
PASTA~\cite{google_multiturn24} applies reinforcement learning to optimize image generation based on user preferences. However, its selection-based approach offers limited fine-grained control over specific object attributes and scene details. 
While these methods have advanced multi-turn T2I generation, only PASTA supports user interaction, though limited to selection. A more comprehensive regeneration strategy is needed—one that extends beyond object positioning, enables automated self-enhancement, and allows fine-grained user control. Our approach fills this gap with an MLLM-driven evaluation agent, which autonomously generates improvement suggestions for iterative self-enhancement while also supporting human-in-the-loop intervention for fine-grained control.

\section{T2I-Copilot: Mutli-Agent T2I System}
\label{sec:method}

T2I-Copilot is a training-free multi-agent system designed to enhance Text-to-Image generation by better interpreting input to enhance generated image quality effectively. As illustrated in Fig.~\ref{fig:pipeline}, T2I-Copilot comprises three sequentially collaborating agents: \textit{Input Interpreter ($A_{in}$)} analyzes and refines user input by clarifying ambiguities and structuring the request into an Analysis Report ($R_A$). \textit{Generation Engine ($A_{gen}$)} selects and executes the most appropriate model based on the task intention and model capabilities. \textit{Quality Evaluator ($A_{eval}$)} assesses generated image quality and text-image alignment, iterating with refinement suggestions if necessary. This system design ensures interpretability, adaptability, and controllability, making it robust against ambiguous inputs and enhancing the overall alignment between user intent and generated images. \textit{For clarity, we provide the pseudocode in the Supplement.} 

\subsection{Input Interpreter Agent ($A_{in}$)}
Users may struggle to craft precise prompts that align with model capabilities or users might not know how models interpret their prompts, leading to unintended outputs. 
To mitigate this, we introduce the Input Interpreter Agent ($A_{in}$), designed to analyze user input—including text prompts and optional reference images—and transform it into a structured Analysis Report ($R_A$), which captures key details essential for model selection and image generation, to facilitate high-quality outputs that better aligns user intent. The agent performs three key functions:
\begin{enumerate}
    \item Input Understanding: Extracts key elements, identifies ambiguities and formulates clarification queries. 
    \item Clarification Resolution: Resolves ambiguities via a MLLM automatic completion or human interaction. 
    \item Summarization: Aggregates the identified key elements and clarified information into a structured Analysis Report ($R_{A}$) for subsequent processing.
\end{enumerate}

\begin{figure}[h]
  \centering
  \includegraphics[width=1.0\linewidth]{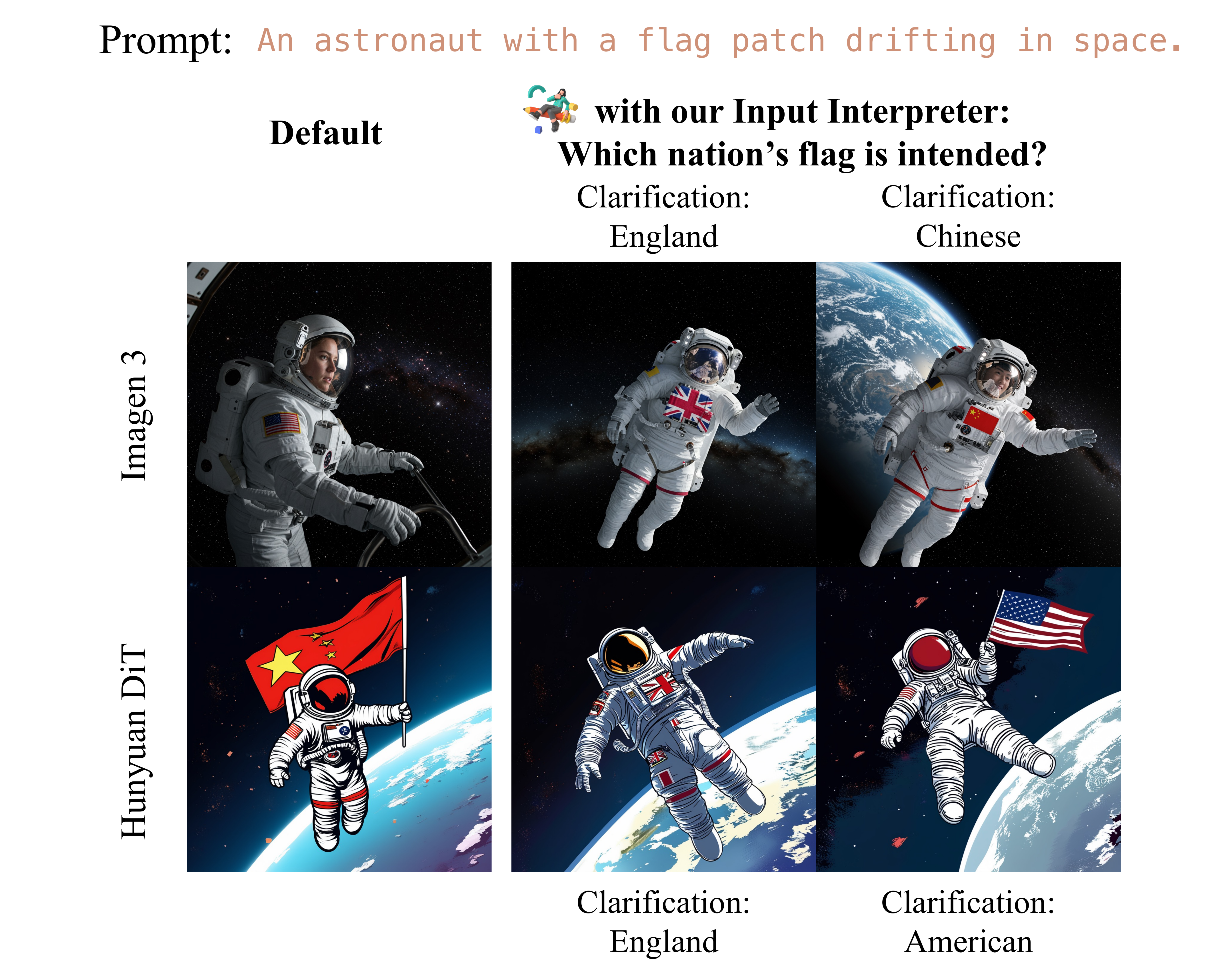}
  \vspace{-0.8cm}
   \caption{\textbf{The effectiveness of Input Interpreter $A_{in}$.} Without clarification, ambiguous terms rely on model-specific knowledge. Our Input Interpreter provides contextual details pre-generation, reducing unintended outputs.
   } 
   \label{fig:input-flag-clarification}
\end{figure}

To achieve this, $A_{in}$ identifies key subjects and attributes within the prompt, analyzing aspects, including background, composition, color harmony, lighting, focus sharpness, emotional impact, uniqueness, creativity, and visual style. It then detects unclear elements, explains potential ambiguities, and generates clarification queries. These are addressed through MLLM reasoning or by requesting user clarification. Take Fig.~\ref{fig:input-flag-clarification} as an example. The prompt ``An astronaut with a flag patch drifting in space'' lacks specificity—which nation's flag is intended? Without clarification, models rely on their default biases (\eg, Imagen 3~\cite{imagen3_24} defaults to a US flag, whereas HunyuanDiT~\cite{hunyuandit_24} defaults to a Chinese flag). By incorporating our Input Interpreter, the model acquires contextual details before generation, reducing unintended outputs.

Beyond ambiguity resolution, the agent dynamically infers details based on user responses and a creativity level parameter $C_{level}$, which controls the extent of automatic enhancement:

\begin{itemize}
    \item \texttt{LOW}: $A_{in}$ strictly adheres to user input. 
    \item \texttt{MEDIUM}: $A_{in}$ makes reasonable assumptions while prioritizing user input.
    \item \texttt{HIGH}: $A_{in}$ autonomously enriches the prompt while ensuring alignment with the user’s original intent when minimal user input is provided.
\end{itemize}

Once analysis is complete, the agent generates an Analysis Report $R_{A}$, structured in JSON format, containing: i) Key extracted elements, attributes, and spatial relationships, ii) Background description, iii) Composition, color harmony, lighting, focus, and style details, iv) User clarifications, and v) Detailed prompt. This report is then passed to the next stage for model selection.

\begin{tcolorbox}[breakable,colback=textboxgrey, colframe=textboxblue, title= Analysis Report, left=1.5mm, right=1.5mm] 
\scriptsize

\texttt{\textbf{Given Prompt}: "A chocolate cupcake with vanilla \\\blank{1cm}frosting on a plate, beside a vanilla cupcake \\\blank{1cm}with chocolate frosting."}\\

\texttt{\textbf{Analysis Report}:}\\
\texttt{\textbf{"Identified elements"}:} \text{\{}\\
\blank{0.3cm}\texttt{\textbf{"main subject"}:} \text{[{\{}}\\
\blank{0.3cm} \textit{\textcolor{commentgreen}{// A list of main objects and corresponding attributes.}} \\
\texttt{\blank{0.6cm}\textbf{"chocolate cupcake"}: "vanilla frosting"}\\ 
\blank{0.6cm} ... \\
\blank{0.3cm}\texttt{\textbf{"references"}:} \text{\{}\\
\blank{0.3cm} \textit{\textcolor{commentgreen}{// Reference images indicating desired content and style.}} \\
\texttt{\textbf{"Creativity fills"}:} \text{\{}\\
\blank{0.3cm} \textit{\textcolor{commentgreen}{// Detailed prompt filled by LLM for critical perspectives.}} \\
\blank{0.3cm} \textit{\textcolor{commentgreen}{// Background, compisition, color harmony, lighting.}} \\
\blank{0.3cm} \textit{\textcolor{commentgreen}{// Focus sharpness, emotional impact, uniqueness creativity, visual style.}} \\
\texttt{\blank{0.3cm}"\textbf{background}": "A simple kitchen table setting to \\\blank{0.6cm}enhance the aesthetic appeal of the cupcakes.",\\
\blank{0.3cm} ... }\\
\texttt{"\textbf{Ambiguous elements}":}  \text{[{\{}}\\
\blank{0.3cm} \textit{\textcolor{commentgreen}{// List the ambiguous elements, reasons, questions for clarification,} \\}
\blank{0.3cm} \textit{\textcolor{commentgreen}{// and user clarification or LLM-generated answer.}} \\

\texttt{\blank{0.6cm}\textbf{"element"}: "plate",\\
        \blank{0.6cm}\textbf{"reason"}: "Type and style of plate are not \\\blank{1cm}specified",\\
        \blank{0.6cm}\textbf{"clarification questions"}: \text{[}\\ 
        \blank{1cm}"What type of plate are you imagining (e.g., \\\blank{1cm}Marble Plate, Plastic Plate)?",\\
        \blank{1cm}"Do you have a preference for the material or \\\blank{1cm}design?" \text{]},\\
        \blank{0.6cm}\textbf{"creativity fill"}: "Assume a simple white \\\blank{1cm}ceramic plate to make it versatile for \\\blank{1cm}presenting desserts"}\text{\},{\{}}\\
       \blank{0.6cm} ...
\end{tcolorbox}

\subsection{Generation Engine Agent ($A_{gen}$)}
Rather than relying on a single state-of-the-art model for T2I generation, we integrate two models to support a diverse range of functionalities. 
These models support both prompt-guided T2I generation and reference-guided T2I editing, allowing fine-grained control over multiple aspects of image synthesis and modification. Although our system utilizes only two models, their complementary capabilities ensure broad coverage across various T2I tasks. 
For generation, the system controls positioning, atmosphere, mood, lighting, and style, ensuring that outputs align closely with user intent. 
For editing, it facilitates object addition, replacement, and removal, offering precise image modifications. 
The Generation Engine selects the most suitable model based on $R_A$ generated by $A_{in}$ and the original user input. If the request originates from a regeneration attempt initiated by the Quality Evaluator, $A_{gen}$ additionally incorporates improvement suggestions and user feedback as input. The workflow consists of three stages: Task Identification, Input Preparation, and Model Execution.

In the \textbf{Task Identification stage}, $A_{gen}$ analyzes the input to determine whether the request involves editing an existing image or generating a new one from scratch. 
This decision is based on both user intent and the capabilities of the available models. For instance, editing models may struggle with certain tasks, such as object rearrangement, style transfer, or fine-grained lighting adjustments. 
If the Quality Evaluator determines that the selected editing model cannot adequately fulfill the request, it suggests switching to the generation model as an alternative. In such cases, $A_{gen}$ reformulates the prompt to effectively approximate the desired modifications, leveraging prompt-based control to achieve the intended outcome. 
This systematic model selection ensures that each request is processed efficiently while maintaining alignment with user intent.

Once the model is selected, $A_{gen}$ enters the \textbf{Input Preparation stage}, refining the input to elicit the model’s full potential and ensure alignment with user intent. 
This process includes generating an optimized prompt that emphasizes critical contents, while adjusting the prompt format to suit the selected model. 
For example, in editing tasks, $A_{gen}$ provides modification-specific descriptions, explicitly specifying object addition, replacement, or removal to ensure precise control. 
When tasks require object-specific modifications, $A_{gen}$ would invoke Referring Expression Segmentation (RES) or prompts the user to annotate target regions via an interactive drawing canvas. This custom-designed tool allows users to mark areas directly on the reference image, enabling a more intuitive and flexible editing experience.
Additionally, if a reference image is provided or if the request involves regeneration, $A_{gen}$ would incorporate the reference to maintain content consistency or guide modifications accordingly.

Finally, in the \textbf{Model Execution stage}, $A_{gen}$ runs the selected model with the prepared inputs to generate the final image $I_{gen}^{n}$, where $n$ is the regeneration count.

\textit{A discussion on the effect of adding more models is in the Supplement.} 

\begin{tcolorbox}[breakable,colback=textboxgrey, colframe=textboxblue, title= Generation Engine, left=1.5mm, right=1.5mm] 
\scriptsize

\texttt{\textbf{"Selected Model"}: "Flux.1-dev",\\
\textbf{"Generating Prompt"}: "A beautifully arranged scene \\\blank{0.3cm}featuring a chocolate cupcake with vanilla \\\blank{0.3cm}frosting on a simple white ceramic plate, \\\blank{0.3cm}alongside a vanilla cupcake with chocolate \\\blank{0.3cm}frosting. The setting is a simple kitchen table, \\\blank{0.3cm}with soft, natural lighting enhancing the textures \\\blank{0.3cm}and colors of the cupcakes. The composition \\\blank{0.3cm}centers the cupcakes at a slight angle, showcasing \\\blank{0.3cm}balance and symmetry, while gentle shadows add \\\blank{0.3cm}depth. The rich brown of the chocolate contrasts \\\blank{0.3cm}nicely with the creamy white of the vanilla \\\blank{0.3cm}frosting, and the vanilla cupcake adds a soft \\\blank{0.3cm}beige tone. The visual style is bright, clean, \\\blank{0.3cm}and fresh, resembling a food blog aesthetic that \\\blank{0.3cm}conveys a sense of indulgence and happiness."\\}
\textit{\textcolor{commentgreen}{// Prompt generated by jointly considering the input \textbf{analysis report}, \textbf{improvement suggestion}, \textbf{user feedback}, and the \textbf{last-round result}.} \\}
\texttt{\textbf{"Reference Content Image"}: None, \\}
\textit{\textcolor{commentgreen}{// Users' input or results from the last iteration,} \\}
\texttt{\textbf{"Reference Style Image"}: None,\\}
\textit{\textcolor{commentgreen}{// Users' input or results from the last iteration,} \\}
\texttt{\textbf{"Reasoning"}: "The prompt requires generating a new \\\blank{0.3cm}scene with detailed aesthetic elements including \\\blank{0.3cm}composition, color harmony, and lighting, which \\\blank{0.3cm}aligns with the strengths of Flux.1-dev. The \\\blank{0.3cm}specific elements and overall atmosphere described \\\blank{0.3cm}suggest a complete new image creation rather than \\\blank{0.3cm}localized edits.",\\}
\textit{\textcolor{commentgreen}{// Reasoning process of model selection.} \\}
\texttt{\textbf{"Condifence Score"}: 0.95
}
\end{tcolorbox}

\subsection{Quality Evaluator Agent ($A_{eval}$)}
Single-turn generation may not always achieve the desired result. To address this, we introduce the Quality Evaluator Agent, which acts as an automatic judge to determine whether the generated image meets user intent. The evaluation is based on two primary criteria: aesthetic quality and text-image alignment. In cases where the output does not fully align with the intended goal, $A_{eval}$ identifies missing elements and provides improvement suggestions. 
Given the generated image, original input, and analysis report, $A_{eval}$ evaluates the image across six sub-fields of aesthetic quality: composition, color harmony, lighting and exposure, focus and sharpness, emotional impact, and uniqueness and creativity. Simultaneously, it examines four sub-fields of text-image alignment: presence of main subjects, accuracy of spatial relationships, adherence to style requirements, and background representation.

If the average score exceeds the predefined \texttt{THRESHOLD}, the generation is complete with no further refinement needed. Conversely, if the score falls below the \texttt{THRESHOLD} or the user requests modifications, $A_{eval}$ redirects the process to $A_{gen}$, incorporating improvement suggestions and user feedback for further refinement, creating an iterative enhancement cycle.

\noindent \textbf{Regeneration request.} 
When regeneration is triggered, $A_{gen}$ re-evaluates model selection while incorporating MLLM-generated improvement suggestions and optional user feedback. This iterative process continues until the output sufficiently aligns with user intent, enhancing the final image quality through progressive refinement. To prevent infinite regeneration loops, we set a termination limit using the hyperparameter \texttt{MAX\_regen\_count}. If the regeneration count reaches this limit, the process stops, and the latest generated image is returned as the final output.

\begin{tcolorbox}[breakable, colback=textboxgrey, colframe=textboxblue, title= Quality Evaluator]
\small
Given the generated image: \\
\includegraphics[width=0.4\textwidth]{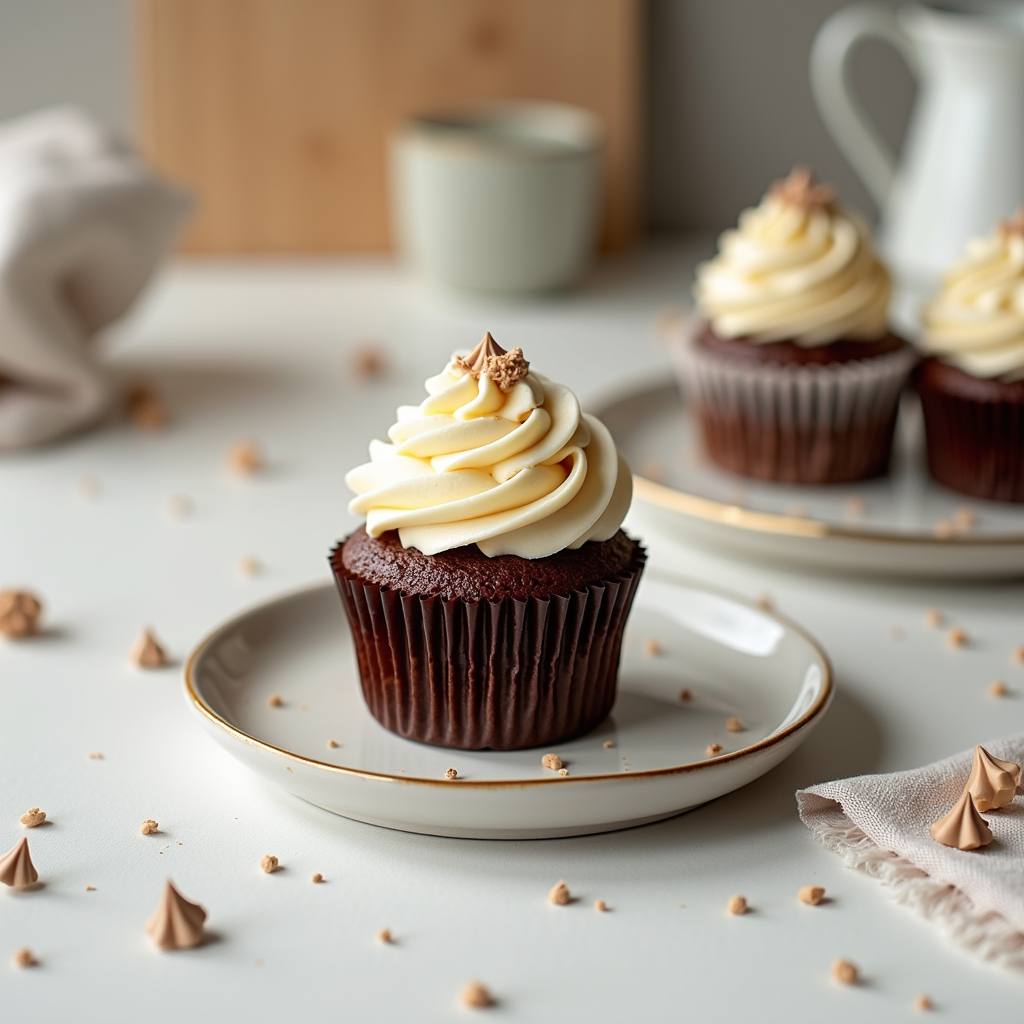}

The evaluation result:\\
\scriptsize
\texttt{\textbf{"Aesthetic Score (0-10)"}}: \text{\{}\\
\textit{\textcolor{commentgreen}{// Score for 6 different aspects, including composition, color harmony, lighting \& exposure, focus \& sharpness, emotional impact, and uniqueness \& creativity.}} \\
\texttt{\blank{0.3cm}\textbf{"Composition"}: 7.5, \\
\blank{0.3cm}\textbf{"Color Harmony"}: 8.5, \\}
\blank{0.6cm}... \\
\texttt{\textbf{"Text-Image Alignment (0-10)"}}: \text{\{}\\
\textit{\textcolor{commentgreen}{// Score for 4 different aspects, including presence of main subjects, accuracy of spatial relationships, adherence to style requirements, and background representation.}} \\
\texttt{\blank{0.3cm}\textbf{"Presence of Main Subjects"}: \textcolor{red}{6.0}, \\
\blank{0.3cm}\textbf{"Accuracy of Spatial Relationships"}: \textcolor{red}{6.5}, \\
\blank{0.6cm}... }\\
\texttt{\textbf{"Missing Elements"}}: \text{[}\\
\texttt{\blank{0.3cm}\textcolor{red}{"Vanilla cupcake with chocolate frosting"},\\
\blank{0.3cm}\textcolor{red}{"Plate arrangement of both cupcakes"}} \text{]},\\
\texttt{\textbf{"Improvement Suggestions"}: "Ensure the vanilla \\\blank{0.3cm}cupcake with chocolate frosting is included in \\\blank{0.3cm}the arrangement, and present both cupcakes on \\\blank{0.3cm}the plate as specified in the prompt."}\\
\texttt{\textbf{"Overall Score"}: 7.65 (< THRESHOLD)}\\

\small
After regeneration, it receives the image shown in Fig.~\ref{fig:pipeline}.

\end{tcolorbox}

\section{Experiments}
\label{sec:exp}

\begin{figure*}[t]
  \centering
  \includegraphics[width=1.0\linewidth]{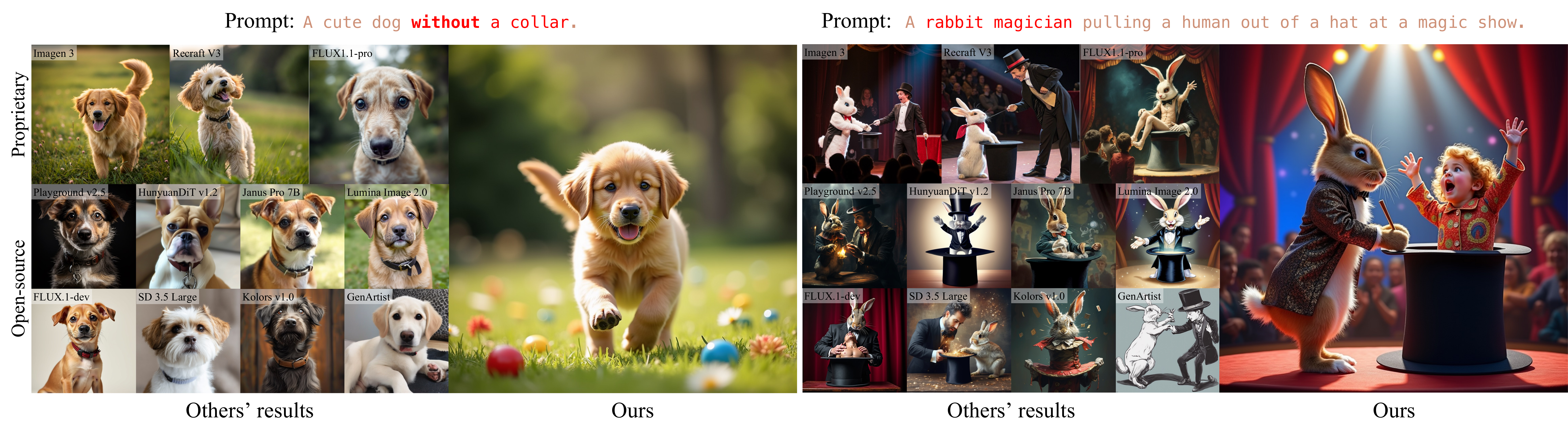}
   \vspace{-0.7cm}
   \caption{\textbf{Qualitative comparison with 11 proprietary and open-source models on two challenging T2I cases.} (Left): Logical negation—only Imagen 3~\cite{imagen3_24} and our method successfully exclude the collar on the dog, while others fail. Our Input Interpreter Agent refines the prompt by explicitly marking the collar as an excluded element, ensuring accurate generation. (Right): Subject-object reversal—only our method and FLUX.1-dev~\cite{flux24} correctly generate a rabbit magician instead of a human magician. Our agent structures the prompt into a detailed analysis report, assigning explicit roles and spatial relationships, enhancing generation accuracy.
   }  
   \label{fig:qualitative}
\end{figure*}

\subsection{Experimental Setup}
\noindent \textbf{Implementation details.}
To ensure a fair comparison, all the reported results from our T2I-Copilot are obtained in automatic mode, without human-in-the-loop, unless otherwise specified. In T2I-Copilot, the prompt-guided T2I generation model is FLUX.1-dev~\cite{flux24}, the reference-guided T2I editing model is PowerPaint~\cite{powerpaint_ECCV24}, (M)LLM is gpt-4o-mini-2024-07-18~\cite{gpt4o_24}, and the Referring Expression Segmentation is Grounding-SAM2~\cite{groundingsam_24}. We set the \texttt{THRESHOLD} as 8.0 and \texttt{MAX\_regen\_count} as 3. 
Our multi-agent system is developed with the framework of LangGraph~\cite{langgraph}.

\noindent \textbf{Baselines.} We compare our proposed method against five proprietary models: Imagen 3 v002~\cite{imagen3_24}, Recraft v3~\cite{recraftv3_24}, FLUX1.1-pro~\cite{flux24}, Midjourney v6~\cite{midjourneyv6.1_24}, and DALLE-3~\cite{dalle3_24}. Additionally, we evaluate it against eight SOTA open-source models: Kolors v1.0~\cite{kolors_arxiv24}, Playground v2.5~\cite{playground_arxiv24}, HunyuanDiT v1.2~\cite{hunyuandit_24}, Janus Pro 7B~\cite{janusPro7B_25}, Lumina Image 2.0~\cite{luminaImage_25}, Stable Diffusion 3.5 Large~\cite{SD35_24}, and FLUX.1-dev~\cite{flux24}. All baselines use official default settings. 
Furthermore, we include an agentic T2I system, GenArtist~\cite{Genartist_NeurIPS24}, with the same controller as ours, \ie, gpt-4o-mini-2024-07-18~\cite{gpt4o_24}. 

\noindent \textbf{Evaluation benchmarks.}  We evaluate model performance on two benchmarks: the widely used DrawBench~\cite{Imagen1_NeurIPS22} and the more challenging GenAI-Bench~\cite{GenAIBench_CVPRW24}. DrawBench consists of 200 samples, while GenAI-Bench contains 1,600 samples, further divided into basic (722 samples) and advanced (871 samples) tasks.\footnote{Seven cases were not categorized as either basic or advanced tasks by the original authors. Upon review, we classified them as basic tasks in our experiments, as each case refers to a single object.} 
Advanced tasks feature complex compositions, including counting, differentiation, comparison, logical negation, and logical universality.

\noindent \textbf{Evaluation metrics.}
We evaluate model performance using the automated metric VQAScore~\cite{VQAScore_ECCV24}, following Imagen 3~\cite{imagen3_24}, which identified it as more human-aligned than CLIPScore~\cite{CLIPScore_EMNLP21}, PickScore~\cite{picks_NeurIPS23}, ImageReward~\cite{imagereward_NeurIPS23}, and HPSv2~\cite{HPS_23}. Furthermore, we conduct a user study to assess both text-image alignment and aesthetic quality.

\begin{figure}[h]
  \centering
  \includegraphics[width=1.0\linewidth]{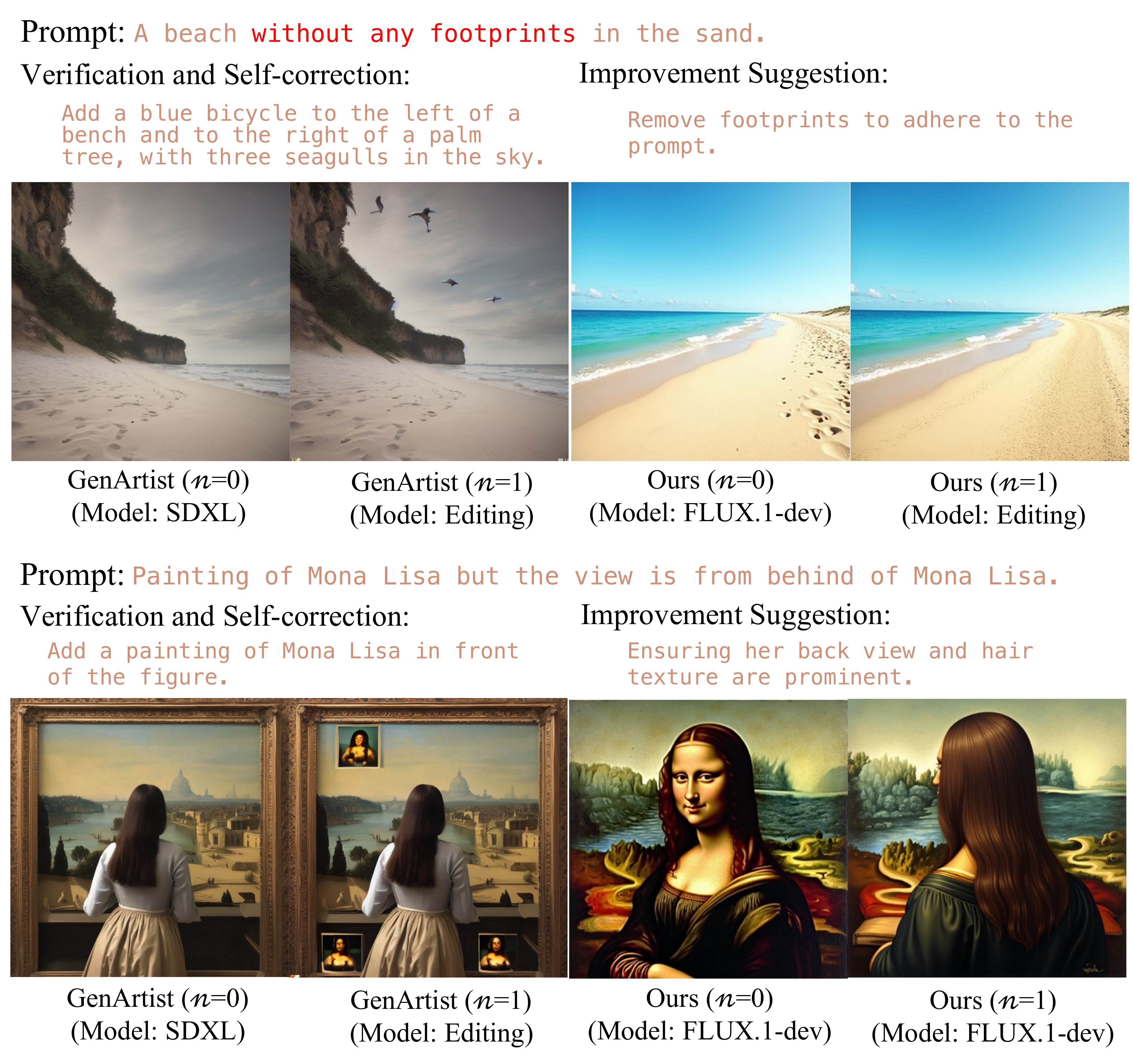}
   \caption{\textbf{The effectiveness of $A_{eval}$: automatic evaluation and regeneration.} Compared to GenArtist~\cite{Genartist_NeurIPS24}, our model provides more contextually relevant suggestions. 
   Ours correctly removes footprints while GenArtist suggests an unrelated fix. For Mona Lisa, ours preserves style and adjusts correctly, while GenArtist loses style and misguides the correction. Both use GPT-4o-mini for evaluation, with $n$ denoting regeneration attempts.}  
   \label{fig:eval-and-regen-cp-Genartist}
\end{figure}

\subsection{Qualitative Results}
We qualitatively compare our results with 11 models, as shown in Fig.~\ref{fig:qualitative}. \textit{More results are provided in Supplement.} The two cases present significant challenges for T2I models. The left case requires logical negation to exclude specified objects from the generated image, while the right case demands precise control over attributes, scene composition, spatial relationships, and action dynamics.

In the left example, only Imagen 3~\cite{imagen3_24} and our method successfully exclude the dog's collar while other models generate it despite its negation. 
Our Input Interpreter Agent ensures this by explicitly marking the collar as excluded in the structured analysis. 
The refined prompt focuses on a ``cute fluffy golden retriever puppy playing outdoors,'' preventing the model from fixating on the negated object and improving prompt adherence. 
In the right example, the challenge lies in the common reversal of subject and object, making it difficult for models to generate a rabbit magician. Only our method and FLUX.1-dev~\cite{flux24} successfully generate the intended concept. Our Input Interpreter Agent resolves this by structuring the prompt into a detailed analysis report, explicitly defining roles and attributes. 
It specifies the rabbit as ``magician outfit, holding a wand, centrally positioned,'' and the human as ``colorful magician's assistant costume, emerging from a hat with a surprised expression.'' 
This structured approach improves generation accuracy and aligns results with user intent.

\noindent \textbf{Performance of Quality Evaluator.} Fig.~\ref{fig:eval-and-regen-cp-Genartist} demonstrates the effectiveness of our evaluation and regeneration quality compared to GenArtist~\cite{Genartist_NeurIPS24}. 
Our model generates suggestions that better align with the original intent of the prompt.
In the first example, when footprints appear despite the prompt specifying a footprint-free beach, our model correctly suggests their removal, whereas GenArtist provides an unrelated correction. 
In the second example, GenArtist generates a back view but loses the Mona Lisa’s style, while our model, though missing the back view, preserves stylistic integrity and provides a more reasonable regeneration suggestion. This improvement stems from our system’s ability to grade images across 10 sub-fields, generating structured improvement suggestions. These suggestions are processed by $A_{eval}$, guiding $A_{gen}$ in selecting the optimal model and preparing suitable input for enhancement.

\subsection{Quantitative Results}

\begin{table*}[t]
    \begin{center}
    \small
    \scalebox{0.76}{%
    \setlength{\tabcolsep}{0.3em}
    \begin{tabular}{lccccccccccccc|cc|c}
        \toprule
        \multirow{4}{*}{Method} & \multicolumn{15}{c}{GenAI-Bench} & \multirow{4}{*}{DrawBench}\\
        \cmidrule(lr){2-16}
        & \multicolumn{6}{c}{Basic} & \multicolumn{6}{c}{Advanced} & \multirow{3}{*}{Overall} & \multicolumn{2}{c}{User Study} & \\
        \cmidrule(lr){2-7}\cmidrule(lr){8-13}\cmidrule(lr){15-16}
        & \multirow{2}{*}{Attribute} & \multirow{2}{*}{Scene} & \multicolumn{3}{c}{Relation} & \multirow{2}{*}{Overall} & \multirow{2}{*}{Count} & \multirow{2}{*}{Differ} & \multirow{2}{*}{Compare} & \multicolumn{2}{c}{Logical} & \multirow{2}{*}{Overall} & & \multirowcell{2}{T2I \\ Alignment} & \multirowcell{2}{Aesthetic \\ Quality} & \\
        \cmidrule(lr){4-6}\cmidrule(lr){11-12}
        & & & Spatial & Action & Part & & & & & Negate & Universal & & & & & \\
        \midrule
        \rowcolor[HTML]{EFEFEF} 
        \textit{Proprietary} & & & & & & & & & & & & & & & & \\
        Imagen 3 v002$^{*}$~\cite{imagen3_24} & 0.909 & \textbf{0.923} & 0.909 & \textbf{0.903} & \textbf{0.918} & 0.912 & \textbf{0.841} & \textbf{0.841} & 0.795 & \textbf{0.673} & 0.788 & \textbf{0.776} & \textbf{0.839} & 95.9\% & 59.9\% & \textbf{0.866}\\
        \scriptsize{(Task completion rate)} & \scriptsize{(92.4\%)} & \scriptsize{(93.1\%)} & \scriptsize{(93.0\%)} & \scriptsize{(89.3\%)} & \scriptsize{(88.7\%)} & \scriptsize{(92.3\%)} & \scriptsize{(91.5\%)} & \scriptsize{(88.9\%)} & \scriptsize{(90.7\%)} & \scriptsize{(88.8\%)} & \scriptsize{(89.1\%)} & \scriptsize{(90.7\%)} & \scriptsize{(91.4\%)} & - & - & \scriptsize{(97.0\%)}\\
        Recraft v3$^{*}$~\cite{recraftv3_24} & \textbf{0.914} & 0.913 & 0.913 & 0.901 & 0.913 & \textbf{0.913} & 0.806 & 0.797 & 0.772 & 0.589 & 0.761 & 0.725 & 0.811 & 89.2\% & 54.1\% & 0.836\\
        FLUX1.1-pro$^{*}$~\cite{flux24} & 0.890 & 0.899 & 0.884 & 0.871 & 0.894 & 0.884 & 0.766 & 0.788 & 0.751 & 0.490 & 0.710 & 0.666 & 0.766 & 95.5\% & 84.2\% & 0.786\\
        Midjourney v6$^{\dagger}$~\cite{midjourneyv6.1_24} & 0.880 & 0.870 & 0.870 & 0.870 & 0.910 & 0.870 & 0.780 & 0.780 & 0.790 & 0.500 & 0.760 & 0.690 & 0.772 & - & - & -\\
        DALL-E 3$^{\dagger}$~\cite{dalle3_24} & 0.910 & 0.900 & \textbf{0.920} & 0.890 & 0.910 & 0.900 & 0.820 & 0.780 & \textbf{0.820} & 0.480 & \textbf{0.800} & 0.700 & 0.791 & - & - & -\\
        \midrule
        \rowcolor[HTML]{EFEFEF} 
        \textit{Open-source} & & & & & & & & & & & & & & & & \\
        Kolors v1.0~\cite{kolors_arxiv24} & 0.821 & 0.841 & 0.832 & 0.818 & 0.803 & 0.819 & 0.737 & 0.726 & 0.705 & 0.438 & 0.695 & 0.621 & 0.711 & 96.8\% & 66.2\% & 0.646 \\
        Playground v2.5~\cite{playground_arxiv24} & 0.818 & 0.850 & 0.803 & 0.818 & 0.821 & 0.815 & 0.732 & 0.696 & 0.721 & 0.499 & 0.695 & 0.640 & 0.720 & 85.8\% & 68.9\% & 0.743 \\
        HunyuanDiT v1.2~\cite{hunyuandit_24} & 0.817 & 0.855 & 0.825 & 0.827 & 0.798 & 0.818 & 0.732 & 0.723 & 0.743 & 0.475 & 0.692 & 0.640 & 0.721 & 94.6\% & 80.2\% & 0.712\\
        Janus Pro-7B~\cite{janusPro7B_25} & 0.865 & 0.886 & 0.867 & 0.856 & 0.870 & 0.859 & 0.731 & 0.759 & 0.734 & 0.480 & 0.693 & 0.653 & 0.747 & 98.2\% & 94.1\% & 0.786\\
        Lumina-Image-2.0~\cite{luminaImage_25} & 0.879 & 0.896 & 0.876 & 0.872 & 0.885 & 0.874 & 0.760 & 0.767 & 0.729 & 0.451 & 0.723 & 0.649 & 0.752 & 99.1\% & 92.8\% & 0.790\\
        SD 3.5 large~\cite{SD35_24} & 0.891 & 0.895 & 0.889 & 0.880 & 0.895 & 0.890 & 0.760 & 0.763 & 0.743 & 0.471 & 0.707 & 0.659 & 0.764 & 92.8\% & 79.3\% & 0.781\\
        FLUX.1-dev~\cite{flux24} & 0.873 & 0.875 & 0.862 & 0.853 & 0.875 & 0.864 & 0.747 & 0.756 & 0.733 & 0.456 & 0.711 & 0.646 & 0.745 & 94.6\% & 86.0\% & 0.769\\
        GenArtist~\cite{Genartist_NeurIPS24} & 0.702 & 0.736 & 0.659 & 0.677 & 0.688 & 0.693 & 0.553 & 0.473 & 0.518 & 0.437 & 0.546 & 0.504 & 0.588 & 97.3\% & 89.2\% & 0.607 \\
        T2I-Copilot (LLM) & 0.893 & 0.909 & 0.893 & 0.885 & 0.899 & 0.892 & 0.813 & 0.807 & 0.759 & 0.659 & 0.766 & 0.747 & 0.813 & - & - & 0.829\\
        T2I-Copilot (Human) & \textbf{0.901} & \textbf{0.917} & \textbf{0.905} & \textbf{0.895} & \textbf{0.902} & \textbf{0.904} & \textbf{0.835} & \textbf{0.820} & \textbf{0.788} & \textbf{0.716} & \textbf{0.798} & \textbf{0.784} & \textbf{0.839} & - & - & \textbf{0.865}\\
        \midrule
        \rowcolor[HTML]{E3F2FD} 
        Max Relative Range & & & & & & & & & & & & & & & & \\
        Proprietary + Open-source & 11.1\% & 9.3\% & 13.4\% & 9.8\% & 13.7\% & 11.2\% & 14.3\% & 19.0\% & 15.3\% & \textcolor{blue}{45.8\%} & 14.8\% & 22.9\% & 16.6\% & - & - & 28.4\% \\
        Open-source & 9.0\% & 7.8\% & 10.8\% & 8.1\% & 12.0\% & 9.2\% & 11.3\% & 15.5\% & 7.7\% & \textcolor{blue}{45.4\%} & 10.8\% & 19.6\% & 13.9\% & - & - & 24.8\% \\
        \bottomrule
    \end{tabular}}
    \end{center}
    \vspace{-0.3cm}
    \footnotesize{$^{*}$Imagen 3 v002~\cite{imagen3_24} results were generated at 23 Feb., 2025. Recraft v3~\cite{recraftv3_24} and FLUX1.1-pro~\cite{flux24} results are generated at 1 Mar., 2025. $^{\dagger}$Midjourney v6~\cite{midjourneyv6.1_24} and DALL-E 3~\cite{dalle3_24} results are from Table 10 in VQAScore~\cite{VQAScore_ECCV24}.}\\
    \vspace{-0.5cm}
    \caption{\textbf{Quantitative comparison of T2I-Copilot against 13 methods on DrawBench~\cite{Imagen1_NeurIPS22} and GenAI-Bench~\cite{GenAIBench_CVPRW24}}, evaluated using VQAScore~\cite{VQAScore_ECCV24}. User study evaluates T2I alignment and aesthetic quality based on win rates. The table also reports Maximum Relative Range, identifying logical negation as the most challenging category. Bold text denotes the best proprietary and open-source models.}
    \label{tab:quan_t&i_alignment}
    \vspace{-0.3cm}
\end{table*}

Tab.~\ref{tab:quan_t&i_alignment} presents a comparative analysis of T2I-Copilot against 13 baselines in terms of VQAScore on DrawBench~\cite{Imagen1_NeurIPS22} and GenAI-Bench~\cite{GenAIBench_CVPRW24}, with performance reported separately for different task categories.

T2I-Copilot outperforms all open-source models across tasks and achieves competitive results against proprietary models. In advanced tasks on GenAI-Bench~\cite{GenAIBench_CVPRW24}, despite being built on FLUX.1-dev~\cite{flux24}, it surpasses its foundation by 15.65\%. This improvement stems from our \textit{Input Interpreter} and \textit{Quality Evaluator} agents, which enhance FLUX.1-dev~\cite{flux24} for better user intent alignment. Against proprietary models, T2I-Copilot outperforms RecraftV3~\cite{recraftv3_24}, FLUX1.1-pro~\cite{flux24}, Midjourney v6~\cite{midjourneyv6.1_24}, and DALLE-3~\cite{dalle3_24} by 3.05\%, 12.09\%, 8.22\%, and 6.68\%, respectively. These results highlight its robustness in handling complex text-image alignment challenges.

Additionally, we computed the Maximum Relative Range (MRR) to measure performance variation across categories, defined as $\frac{\max(X) - \min(X)}{\text{mean}(X)} \times 100\%$, where $X$ represents the performance scores. A lower MRR indicates more consistent performance within a category. 
Excluding our method with humans and one outlier, Table~\ref{tab:quan_t&i_alignment} shows the highest MRR in \textit{logical negation}, a challenging task requiring models to exclude specified objects, demanding strong reasoning (e.g., left sample in Fig.~\ref{fig:qualitative}, top sample in Fig.~\ref{fig:eval-and-regen-cp-Genartist}). 
Our system tackles logical negation with the \textit{Input Interpreter} agent, leveraging LLM reasoning to enhance comprehension and text-to-image alignment. Among open-source models, it outperforms all competitors by at least 31.95\%. 
The second-best, Playground v2.5~\cite{playground_arxiv24}, improves prompt adherence through fine-tuning, while the third-best, Janus-pro-7B~\cite{janusPro7B_25}, enhances comprehension via a training-phase module. Without relying on finetuning, our approach uses explicit prompt reasoning during inference, ensuring adaptability. 
Compared to proprietary models, our method outperforms RecraftV3~\cite{recraftv3_24}, FLUX1.1-pro~\cite{flux24}, Midjourney v6~\cite{midjourneyv6.1_24}, and DALL-E 3~\cite{dalle3_24} by at least 11.8\%, achieving competitive performance with Imagen 3~\cite{imagen3_24}. These results highlight the effectiveness of our approach in tackling complex text-image alignment, particularly those requiring logical reasoning and precise prompt comprehension.

\noindent \textbf{Human-in-the-loop.} 
In Tab.~\ref{tab:quan_t&i_alignment}, we further incorporate human feedback into the \textit{Quality Evaluator} Agent to compare enhancement directions identified by humans and LLMs. Human input improves text-image alignment in the VQAScore by an additional 3.17\% across the GenAI-Bench dataset. This demonstrates that integrating human interaction into the system enhances control and better aligns outputs with human intent.

\noindent \textbf{User Study.} 
In Tab.~\ref{tab:quan_t&i_alignment}, we present a user study on text-image alignment and aesthetic quality. We randomly sampled 33 image sets, each method contributing three samples, totaling 2,442 votes. 
Each comparison included an image from our method and one from a baseline. For each set, volunteers answered two questions: (1) selecting the image that best aligned with the text prompt and (2) choosing the one they found more visually appealing beyond text alignment.
Our method achieved an average win rate of 94.5\% for text-image alignment and 77.7\% for aesthetic quality, suggesting that participants placed greater emphasis on factors like composition and style when evaluating visual appeal. While alignment played a role in perception, aesthetic preferences appeared more subjective. We plan further studies to better understand these factors and refine aesthetic quality beyond text-image alignment.

\noindent \textbf{Ablation study.} 
In Tab.~\ref{tab:ablation}, we conduct the ablation study to evaluate the impact of the proposed \textit{Input Interpreter} and \textit{Quality Evaluator} on GenAI-Bench~\cite{GenAIBench_CVPRW24}. 
The results show that $A_{in}$ and $A_{eval}$ contribute 7.69\% and 0.92\% improvements, respectively, in text-to-image alignment. This demonstrates that effectively interpreting the input plays a crucial role in enhancing image generation quality.

\begin{table}[h]
    \begin{center}
    \scalebox{.8}{%
    \setlength{\tabcolsep}{0.3em}
    \begin{tabular}{lccc}
        \toprule
        Method & Basic & Advanced & All\\
        \midrule
        $A_{gen}$ + $A_{eval}$ (w/o $A_{in}$) & 0.864 & 0.646 & 0.755 \\
        $A_{in}$ + $A_{gen}$ (w/o $A_{eval}$) & 0.888 & 0.736 & 0.805\\
        $A_{in}$ + $A_{gen}$ + $A_{eval}$ & 0.892 & 0.747 & 0.813\\
        \bottomrule
    \end{tabular}}
    \end{center}
    \vspace{-0.5cm}
    \caption{\textbf{Ablation study on GenAI-Bench~\cite{GenAIBench_CVPRW24}.}}
    \label{tab:ablation}
    \vspace{-0.6cm}
\end{table}

\section{Conclusion}
\label{sec:conclusion}

In this work, we introduced T2I-Copilot, a training-free multi-agent system designed to enhance interpretability, controllability, and efficiency in Text-to-Image generation. By integrating three specialized agents—Input Interpreter, Generation Engine, and Quality Evaluator—our approach addresses key challenges in prompt interpretation, model selection, and iterative refinement. Without relying on fine-tuning or architectural modifications, T2I-Copilot operates autonomously while incorporating human-in-the-loop interaction, ensuring adaptability across diverse prompts and user needs. Our evaluation on GenAI-Bench demonstrates that T2I-Copilot achieves a VQAScore comparable to Recraft V3 and Imagen 3, surpasses FLUX1.1-pro by 6.17\% at only 12.48\% of its cost, and outperforms FLUX.1-dev and SD 3.5 Large by 9.11\% and 6.36\%, respectively. 
\section{Acknowledgments}
\label{sec:ack}

This research was supported in part by National Science Foundation under Award \#2427478 - CAREER Program, and by National Science Foundation and the Institute of Education Sciences, U.S. Department of Education under Award \#2229873 - National AI Institute for Exceptional Education. This project was also partially supported by cyberinfrastructure resources and services provided by College of Computing at the Georgia Institute of Technology, Atlanta, Georgia, USA. We sincerely thank Fengzhe Zhou for valuable suggestions and Teng-Fang Hsiao for insights on the editing model. We also appreciate Ali Hassani, Kai Wang and Aditya Kane for their kind support on server logistics.

{
    \small
    \bibliographystyle{ieeenat_fullname}
    \bibliography{main}

\begin{thebibliography}{44}
\providecommand{\natexlab}[1]{#1}
\providecommand{\url}[1]{\texttt{#1}}
\expandafter\ifx\csname urlstyle\endcsname\relax
  \providecommand{\doi}[1]{doi: #1}\else
  \providecommand{\doi}{doi: \begingroup \urlstyle{rm}\Url}\fi

\bibitem[AI(2024)]{SD35_24}
Stability AI.
\newblock Stable diffusion 3.5, 2024.

\bibitem[Bai et~al.(2025)Bai, Chen, Liu, Wang, Ge, Song, Dang, Wang, Wang, Tang, Zhong, Zhu, Yang, Li, Wan, Wang, Ding, Fu, Xu, Ye, Zhang, Xie, Cheng, Zhang, Yang, Xu, and Lin]{QWen2.5_VL}
Shuai Bai, Keqin Chen, Xuejing Liu, Jialin Wang, Wenbin Ge, Sibo Song, Kai Dang, Peng Wang, Shijie Wang, Jun Tang, Humen Zhong, Yuanzhi Zhu, Mingkun Yang, Zhaohai Li, Jianqiang Wan, Pengfei Wang, Wei Ding, Zheren Fu, Yiheng Xu, Jiabo Ye, Xi Zhang, Tianbao Xie, Zesen Cheng, Hang Zhang, Zhibo Yang, Haiyang Xu, and Junyang Lin.
\newblock Qwen2.5-vl technical report.
\newblock \emph{arXiv preprint arXiv:2502.13923}, 2025.

\bibitem[Baldridge et~al.(2024)Baldridge, Bauer, Bhutani, Brichtova, Bunner, and et~al.]{imagen3_24}
Jason Baldridge, Jakob Bauer, Mukul Bhutani, Nicole Brichtova, Andrew Bunner, and Kelvin~Chan et al.
\newblock Imagen 3.
\newblock \emph{arXiv preprint arXiv:2408.07009}, 2024.

\bibitem[Chefer et~al.(2023)Chefer, Alaluf, Vinker, Wolf, and Cohen-Or]{A&E_siggraph23}
Hila Chefer, Yuval Alaluf, Yael Vinker, Lior Wolf, and Daniel Cohen-Or.
\newblock Attend-and-excite: Attention-based semantic guidance for text-to-image diffusion models.
\newblock In \emph{ACM Special Interest Group on Computer Graphics and Interactive Techniques (SIGGRAPH)}, 2023.

\bibitem[Chen et~al.(2024{\natexlab{a}})Chen, Tseng, Tsao, and Shuai]{CAT_NeurIPS24}
Chieh{-}Yun Chen, Chiang Tseng, Li{-}Wu Tsao, and Hong{-}Han Shuai.
\newblock A cat is {A} cat (not {A} dog!): Unraveling information mix-ups in text-to-image encoders through causal analysis and embedding optimization.
\newblock In \emph{Advances in Neural Information Processing Systems (NeurIPS)}, 2024{\natexlab{a}}.

\bibitem[Chen et~al.(2025)Chen, Wu, Liu, Pan, Liu, Xie, Yu, and Ruan]{janusPro7B_25}
Xiaokang Chen, Zhiyu Wu, Xingchao Liu, Zizheng Pan, Wen Liu, Zhenda Xie, Xingkai Yu, and Chong Ruan.
\newblock Janus-pro: Unified multimodal understanding and generation with data and model scaling.
\newblock \emph{arXiv preprint arXiv:2501.17811}, 2025.

\bibitem[Chen et~al.(2024{\natexlab{b}})Chen, Li, Wang, Chen, Jiang, Li, Wang, Yang, and Tai]{RAG_arxiv24}
Zhennan Chen, Yajie Li, Haofan Wang, Zhibo Chen, Zhengkai Jiang, Jun Li, Qian Wang, Jian Yang, and Ying Tai.
\newblock Region-aware text-to-image generation via hard binding and soft refinement.
\newblock \emph{arXiv preprint arXiv:2411.06558}, 2024{\natexlab{b}}.

\bibitem[Hao et~al.(2023)Hao, Chi, Dong, and Wei]{Po_NeurIPS23}
Yaru Hao, Zewen Chi, Li Dong, and Furu Wei.
\newblock Optimizing prompts for text-to-image generation.
\newblock In \emph{Advances in Neural Information Processing Systems (NeurIPS)}, 2023.

\bibitem[Hessel et~al.(2021)Hessel, Holtzman, Forbes, Bras, and Choi]{CLIPScore_EMNLP21}
Jack Hessel, Ari Holtzman, Maxwell Forbes, Ronan~Le Bras, and Yejin Choi.
\newblock Clipscore: {A} reference-free evaluation metric for image captioning.
\newblock In \emph{Proceedings of the Conference on Empirical Methods in Natural Language Processing (EMNLP)}, pages 7514--7528, 2021.

\bibitem[Hu et~al.(2024)Hu, Li, van~de Weijer, Gao, Khan, Yang, Cheng, Wang, and Wang]{tome_NeurIPS24}
Taihang Hu, Linxuan Li, Joost van~de Weijer, Hongcheng Gao, Fahad~Shahbaz Khan, Jian Yang, Ming{-}Ming Cheng, Kai Wang, and Yaxing Wang.
\newblock Token merging for training-free semantic binding in text-to-image ynthesis.
\newblock In \emph{Advances in Neural Information Processing Systems (NeurIPS)}, 2024.

\bibitem[Kirstain et~al.(2023)Kirstain, Polyak, Singer, Matiana, Penna, and Levy]{picks_NeurIPS23}
Yuval Kirstain, Adam Polyak, Uriel Singer, Shahbuland Matiana, Joe Penna, and Omer Levy.
\newblock Pick-a-pic: An open dataset of user preferences for text-to-image generation.
\newblock In \emph{Advances in Neural Information Processing Systems (NeurIPS)}, 2023.

\bibitem[Labs(2024)]{flux24}
Black~Forest Labs.
\newblock {FLUX}, 2024.

\bibitem[Li et~al.(2024{\natexlab{a}})Li, Lin, Pathak, Li, Xia, Neubig, Zhang, and Ramanan]{GenAIBench_CVPRW24}
Baiqi Li, Zhiqiu Lin, Deepak Pathak, Jiayao~Emily Li, Xide Xia, Graham Neubig, Pengchuan Zhang, and Deva Ramanan.
\newblock Gen{AI}-bench: A holistic benchmark for compositional text-to-visual generation.
\newblock In \emph{Synthetic Data for Computer Vision Workshop @ CVPR}, 2024{\natexlab{a}}.

\bibitem[Li et~al.(2024{\natexlab{b}})Li, Kamko, Akhgari, Sabet, Xu, and Doshi]{playground_arxiv24}
Daiqing Li, Aleks Kamko, Ehsan Akhgari, Ali Sabet, Linmiao Xu, and Suhail Doshi.
\newblock Playground v2.5: Three insights towards enhancing aesthetic quality in text-to-image generation.
\newblock \emph{arXiv preprint arXiv:2402.17245}, 2024{\natexlab{b}}.

\bibitem[Li et~al.(2024{\natexlab{c}})Li, Zhang, Lin, Xiong, Long, Deng, and et~al.]{hunyuandit_24}
Zhimin Li, Jianwei Zhang, Qin Lin, Jiangfeng Xiong, Yanxin Long, Xinchi Deng, and Yingfang~Zhang et al.
\newblock Hunyuan-{DiT}: A powerful multi-resolution diffusion transformer with fine-grained chinese understanding.
\newblock \emph{arXiv preprint arXiv:2405.08748}, 2024{\natexlab{c}}.

\bibitem[Lian et~al.(2024)Lian, Li, Yala, and Darrell]{LMD_TMLR24}
Long Lian, Boyi Li, Adam Yala, and Trevor Darrell.
\newblock Llm-grounded diffusion: Enhancing prompt understanding of text-to-image diffusion models with large language models.
\newblock \emph{Transactions on Machine Learning Research (TMLR)}, 2024.

\bibitem[Lin et~al.(2024)Lin, Pathak, Li, Li, Xia, Neubig, Zhang, and Ramanan]{VQAScore_ECCV24}
Zhiqiu Lin, Deepak Pathak, Baiqi Li, Jiayao Li, Xide Xia, Graham Neubig, Pengchuan Zhang, and Deva Ramanan.
\newblock Evaluating text-to-visual generation with image-to-text generation.
\newblock In \emph{Proceedings of the European Conference on Computer Vision (ECCV)}, pages 366--384, 2024.

\bibitem[Ma{\~n}as et~al.(2024)Ma{\~n}as, Astolfi, Hall, Ross, Urbanek, Williams, Agrawal, Romero-Soriano, and Drozdzal]{metaPO_TMLR24}
Oscar Ma{\~n}as, Pietro Astolfi, Melissa Hall, Candace Ross, Jack Urbanek, Adina Williams, Aishwarya Agrawal, Adriana Romero-Soriano, and Michal Drozdzal.
\newblock Improving text-to-image consistency via automatic prompt optimization.
\newblock \emph{Transactions on Machine Learning Research (TMLR)}, 2024.

\bibitem[Midjourney(2024)]{midjourneyv6.1_24}
Midjourney.
\newblock {Midjourney v6.1}, 2024.

\bibitem[{Mistral AI}(2025)]{Mistral_small}
{Mistral AI}.
\newblock {Mistral Small 3.1 24B}, 2025.

\bibitem[Nabati et~al.(2024)Nabati, Tennenholtz, Hsu, Ryu, Ramachandran, Chow, Li, and Boutilier]{google_multiturn24}
Ofir Nabati, Guy Tennenholtz, Chih{-}Wei Hsu, Moonkyung Ryu, Deepak Ramachandran, Yinlam Chow, Xiang Li, and Craig Boutilier.
\newblock Personalized and sequential text-to-image generation.
\newblock \emph{arXiv preprint arXiv:2412.10419}, 2024.

\bibitem[OpenAI(2024{\natexlab{a}})]{dalle3_24}
OpenAI.
\newblock {DALL·E} 3, 2024{\natexlab{a}}.

\bibitem[OpenAI(2024{\natexlab{b}})]{gpt4o_24}
OpenAI.
\newblock {GPT}-4o, 2024{\natexlab{b}}.

\bibitem[Podell et~al.(2024)Podell, English, Lacey, Blattmann, Dockhorn, M{\"u}ller, Penna, and Rombach]{sdxl_ICLR24}
Dustin Podell, Zion English, Kyle Lacey, Andreas Blattmann, Tim Dockhorn, Jonas M{\"u}ller, Joe Penna, and Robin Rombach.
\newblock {SDXL}: Improving latent diffusion models for high-resolution image synthesis.
\newblock In \emph{Proceedings of the International Conference on Learning Representations (ICLR)}, 2024.

\bibitem[Qin et~al.(2024)Qin, Wu, Chen, Ren, Li, Wu, Xiao, Wang, and Wen]{diffusionGPT_24arxiv}
Jie Qin, Jie Wu, Weifeng Chen, Yuxi Ren, Huixia Li, Hefeng Wu, Xuefeng Xiao, Rui Wang, and Shilei Wen.
\newblock {DiffusionGPT}: Llm-driven text-to-image generation system.
\newblock \emph{arXiv preprint arXiv:2401.10061}, 2024.

\bibitem[Rassin et~al.(2023)Rassin, Hirsch, Glickman, Ravfogel, Goldberg, and Chechik]{syngen_nips23}
Royi Rassin, Eran Hirsch, Daniel Glickman, Shauli Ravfogel, Yoav Goldberg, and Gal Chechik.
\newblock Linguistic binding in diffusion models: Enhancing attribute correspondence through attention map alignment.
\newblock In \emph{Advances in Neural Information Processing Systems (NeurIPS)}, 2023.

\bibitem[Recraft(2024)]{recraftv3_24}
Recraft.
\newblock {Recraft v3}, 2024.

\bibitem[Ren et~al.(2024)Ren, Liu, Zeng, Lin, Li, Cao, Chen, Huang, Chen, Yan, Zeng, Zhang, Li, Yang, Li, Jiang, and Zhang]{groundingsam_24}
Tianhe Ren, Shilong Liu, Ailing Zeng, Jing Lin, Kunchang Li, He Cao, Jiayu Chen, Xinyu Huang, Yukang Chen, Feng Yan, Zhaoyang Zeng, Hao Zhang, Feng Li, Jie Yang, Hongyang Li, Qing Jiang, and Lei Zhang.
\newblock Grounded sam: Assembling open-world models for diverse visual tasks.
\newblock \emph{arXiv preprint arXiv:2401.14159}, 2024.

\bibitem[Saharia et~al.(2022)Saharia, Chan, Saxena, Li, Whang, Denton, Ghasemipour, Lopes, Ayan, Salimans, Ho, Fleet, and Norouzi]{Imagen1_NeurIPS22}
Chitwan Saharia, William Chan, Saurabh Saxena, Lala Li, Jay Whang, Emily~L. Denton, Seyed Kamyar~Seyed Ghasemipour, Raphael~Gontijo Lopes, Burcu~Karagol Ayan, Tim Salimans, Jonathan Ho, David~J. Fleet, and Mohammad Norouzi.
\newblock Photorealistic text-to-image diffusion models with deep language understanding.
\newblock In \emph{Advances in Neural Information Processing Systems (NeurIPS)}, 2022.

\bibitem[Schmidgall et~al.(2025)Schmidgall, Su, Wang, Sun, Wu, Yu, Liu, Liu, and Barsoum]{agentLab_25}
Samuel Schmidgall, Yusheng Su, Ze Wang, Ximeng Sun, Jialian Wu, Xiaodong Yu, Jiang Liu, Zicheng Liu, and Emad Barsoum.
\newblock Agent laboratory: Using {LLM} agents as research assistants.
\newblock \emph{arXiv preprint arXiv:2501.04227}, 2025.

\bibitem[Shen et~al.(2023)Shen, Song, Tan, Li, Lu, and Zhuang]{huggingGPT_NeurIPS23}
Yongliang Shen, Kaitao Song, Xu Tan, Dongsheng Li, Weiming Lu, and Yueting Zhuang.
\newblock Hugginggpt: Solving {AI} tasks with chatgpt and its friends in huggingface.
\newblock In \emph{Advances in Neural Information Processing Systems (NeurIPS)}, 2023.

\bibitem[Team(2024)]{kolors_arxiv24}
Kolors Team.
\newblock Kolors: Effective training of diffusion model for photorealistic text-to-image synthesis.
\newblock 2024.

\bibitem[team(2024)]{langgraph}
LangChain team.
\newblock {LangGraph}, 2024.

\bibitem[Team(2025)]{luminaImage_25}
Lumina Team.
\newblock Lumina-image 2.0 : A unified and efficient image generative model, 2025.

\bibitem[Team(2024)]{omost_24}
Omost Team.
\newblock Omost github page (https://github.com/lllyasviel/omost), 2024.

\bibitem[Wang et~al.(2024)Wang, Li, Li, and Liu]{Genartist_NeurIPS24}
Zhenyu Wang, Aoxue Li, Zhenguo Li, and Xihui Liu.
\newblock Genartist: Multimodal {LLM} as an agent for unified image generation and editing.
\newblock In \emph{Advances in Neural Information Processing Systems (NeurIPS)}, 2024.

\bibitem[Wu et~al.(2023{\natexlab{a}})Wu, Yin, Qi, Wang, Tang, and Duan]{visualchatgpt_23}
Chenfei Wu, Shengming Yin, Weizhen Qi, Xiaodong Wang, Zecheng Tang, and Nan Duan.
\newblock Visual chatgpt: Talking, drawing and editing with visual foundation models.
\newblock \emph{arXiv preprint arXiv:2303.04671}, 2023{\natexlab{a}}.

\bibitem[Wu et~al.(2024)Wu, Lian, Gonzalez, Li, and Darrell]{SLD_CVPR24}
Tsung-Han Wu, Long Lian, Joseph~E. Gonzalez, Boyi Li, and Trevor Darrell.
\newblock { Self-Correcting LLM-Controlled Diffusion Models }.
\newblock In \emph{Proceedings of the IEEE/CVF Conference on Computer Vision and Pattern Recognition (CVPR)}, 2024.

\bibitem[Wu et~al.(2023{\natexlab{b}})Wu, Hao, Sun, Chen, Zhu, Zhao, and Li]{HPS_23}
Xiaoshi Wu, Yiming Hao, Keqiang Sun, Yixiong Chen, Feng Zhu, Rui Zhao, and Hongsheng Li.
\newblock Human preference score v2: A solid benchmark for evaluating human preferences of text-to-image synthesis.
\newblock \emph{arXiv preprint arXiv:2306.09341}, 2023{\natexlab{b}}.

\bibitem[Xu et~al.(2023)Xu, Liu, Wu, Tong, Li, Ding, Tang, and Dong]{imagereward_NeurIPS23}
Jiazheng Xu, Xiao Liu, Yuchen Wu, Yuxuan Tong, Qinkai Li, Ming Ding, Jie Tang, and Yuxiao Dong.
\newblock Imagereward: Learning and evaluating human preferences for text-to-image generation.
\newblock In \emph{Advances in Neural Information Processing Systems (NeurIPS)}, 2023.

\bibitem[Yao et~al.(2023)Yao, Zhao, Yu, Du, Shafran, Narasimhan, and Cao]{react_ICLR23}
Shunyu Yao, Jeffrey Zhao, Dian Yu, Nan Du, Izhak Shafran, Karthik~R. Narasimhan, and Yuan Cao.
\newblock React: Synergizing reasoning and acting in language models.
\newblock In \emph{Proceedings of the International Conference on Learning Representations (ICLR)}, 2023.

\bibitem[Zhang et~al.(2024)Zhang, Sohn, Hahn, Shi, and Essa]{finestyle_NeurIPS24}
Gong Zhang, Kihyuk Sohn, Meera Hahn, Humphrey Shi, and Irfan Essa.
\newblock Finestyle: Fine-grained controllable style personalization for text-to-image models.
\newblock In \emph{Advances in Neural Information Processing Systems (NeurIPS)}, 2024.

\bibitem[Zhou et~al.(2024)Zhou, Shao, Bai, Xu, Han, and Xie]{goldennoise_24arxiv}
Zikai Zhou, Shitong Shao, Lichen Bai, Zhiqiang Xu, Bo Han, and Zeke Xie.
\newblock Golden noise for diffusion models: {A} learning framework.
\newblock \emph{arXiv preprint arXiv:2411.09502}, 2024.

\bibitem[Zhuang et~al.(2024)Zhuang, Zeng, Liu, Yuan, and Chen]{powerpaint_ECCV24}
Junhao Zhuang, Yanhong Zeng, Wenran Liu, Chun Yuan, and Kai Chen.
\newblock A task is worth one word: Learning with task prompts for high-quality versatile image inpainting.
\newblock In \emph{Proceedings of the European Conference on Computer Vision (ECCV)}, 2024.

\end{thebibliography}
}

\clearpage

\appendix

{\Large{Supplementary Material\\}}

This supplement includes pseudocode (Supp.~\ref{sec:pseudo-code}), the effect of including more models (Supp.~\ref{sec:more-model}), an ablation study of MLLMs (Supp.~\ref{sec:MLLM-ablation}), cost comparison (Supp.~\ref{sec:cost_comparison}) and more qualitative results (Supp.~\ref{sec:qualitative-results}), covering ambiguous terms, single-turn results, and multi-turn results in both automatic mode and human-in-the-loop settings.

\section{Pseudocode for T2I-Copilot}
\label{sec:pseudo-code}

\begin{algorithm}
\caption{: T2I-Copilot Multi-Agent System}
\label{alg:T2I-Copilot}
\scriptsize
\begin{algorithmic}[1]
\Statex \hspace{-\algorithmicindent} \textbf{Input:} User input prompt $P$, Optional reference image $I_{ref}$, Creativity level $C_{level}$, Human-in-the-loop flag $H_{l}$, Improvement suggestions $S_{imp}$

\State $R_A \gets A_{in}(P, I_{ref}, C_{level}, H_{l})$ \Comment{Generate Analysis Report}
\State $I_{gen}^{n} \gets A_{gen}(R_A, P, I_{ref})$  \Comment{Generate Initial Image}
\State $score, S_{imp} \gets A_{eval}(I_{gen}^{n}, P, R_A)$ \Comment{Evaluate Image Quality}
\State $n \gets 0$
\State $U_f \gets None$ \Comment{Initialize user feedback variable}
\While{$score \textless  \texttt{THRESHOLD}$ \textbf{and} $n \textless \texttt{MAX\_regen\_count}$}
    \If{$H_{l}$}
        \State $U_{f} \gets$ Get user feedback on $I_{gen}^{n}$
    \EndIf
    \State $I_{gen}^{n} \gets A_{gen}(R_A, P, I_{ref}, S_{imp}, U_{f})$ \Comment{Regenerate Image}
    \State $score, S_{imp} \gets A_{eval}(I_{gen}^{n}, P, R_A)$ \Comment{Re-evaluate Image Quality}
    \State $n \gets n + 1$
\EndWhile

\Statex \hspace{-\algorithmicindent} \textbf{return} $I_{gen}^{n}$ \Comment{Final Output Image}

\end{algorithmic}
\end{algorithm}

Our system includes modular error handlers and fallbacks. Failures fall into two categories, both handled automatically.
First, for region extraction errors, RES segments unwanted objects or MLLM generates masks from bounding boxes; if both fail, MLLM infers boxes from prompt-image context, and RES is retried with full prompts. Only after all options fail is an error raised in $A_{\text{gen}}$, triggering fallback to the previous image or notifying the user (0.1\% of cases).
Second, for format extraction errors, malformed outputs trigger an automatic MLLM retry, which typically succeeds (0.3\% for GPT-4o-mini, 1\% for Qwen2.5-VL-3B).
These mechanisms localize errors and prevent cascading failures.

\section{The effect of including more models}
\label{sec:more-model}
Our experiments show that the effect varies depending on specific conditions. We initially incorporated five models for selection, including a position-aware T2I model, RAG-Diffusion~\cite{RAG_arxiv24}, a reference-based IP-Adapter, and a reference-based style transfer model. However, we found that the last two models were rarely selected. Moreover, using RAG-Diffusion led to a 3.43\% decrease in performance on VQAScore~\cite{VQAScore_ECCV24} on GenAI-Bench~\cite{GenAIBench_CVPRW24}, and the performance of the position-aware model was inconsistent. This inconsistency stemmed from the dependency on LLMs for position separation and RAG-Diffusion’s effectiveness in following the designed positional relationships. Instead, we found that simple reprompting in a prompt-guided T2I model could achieve similar positional control without the added complexity.

Similarly, integrating IP-Adapter did not significantly improve adherence to reference images for reference-based generation, leading to a slight performance drop of 0.39\%. Instead, we could directly use a reference-based editing model to incorporate this functionality more effectively.

Furthermore, including more models requires careful system prompt design for model selection. Without proper prompt tuning, many tools remain unused. For instance, GenArtist~\cite{Genartist_NeurIPS24} includes 10 models for T2I generation and 8 for editing. The default super-resolution tool is excluded from the selection process because it is directly applied to every generated sample rather than being chosen dynamically. As a result, among the remaining models, only 3 generation models and 2 editing models were selected when generating 1,800 images in DrawBench~\cite{Imagen1_NeurIPS22} and GenAI-Bench~\cite{GenAIBench_CVPRW24}, leaving 6 generation tools and 6 editing tools unused. This highlights the need to evaluate whether adding more models meaningfully contributes to performance improvements.

\renewcommand{\thetable}{A}
\begin{table*}[t]
    \begin{center}
    \small
    \resizebox{\textwidth}{!}{%
    \setlength{\tabcolsep}{0.3em}
    \begin{tabular}{lccc|ccccc|ccr}
        \toprule
        \multirowcell{2}[0pt][l]{Method  {\scriptsize (MLLM backbone)}} & \multicolumn{3}{c|}{Performance (VQAScore)} & \multicolumn{4}{c}{Inference time (s)} & \multirowcell{2}{Multi-turn\\ Turns} & \multicolumn{3}{c}{Cost/image ( $10^{-3}$ USD)} \\
        \cmidrule(lr){2-4}\cmidrule(lr){5-8}\cmidrule(lr){10-12}
        & Basic & Advanced & Overall & MLLM Latency & Generator & Editor & End-to-End & & LLM & T2I & All\\
        \midrule
        FLUX1.1-pro & 0.884 & 0.666 & 0.766 & - & - & - & 3.38 & 1.0 & - & { \scriptsize $\text{40.0}^\text{API}$} & 40.0 \\
        GenArtist {\scriptsize (GPT-4o-mini)} & 0.693 & 0.504 & 0.588 & 5.75 & 5.7& 3.7 & 56.96 & 3.3 & {\scriptsize $\text{3.1}^\text{API}$} & 1.6 & 4.7\\
        Ours  {\scriptsize (GPT-4o-mini)} & 0.892 & 0.747 & 0.813 & 23.7 & 17.4 & 5.9 & 45.72 & 1.5 & {\scriptsize $\text{5.0}^\text{API}$} & 1.6 & 6.6\\
        Ours {\scriptsize (Mistral Small 3.1-24B)} & 0.893 & 0.761 & 0.821 & 44.0 & 17.6 & 5.7 & 64.74 & 1.5 & 7.3 & 1.5 & 8.9 \\
        Ours {\scriptsize (Qwen2.5-VL-7B)} & 0.893 & 0.743 & 0.811 & 26.9 &17.9 & 6.2 & 57.24 & 1.6 & 2.2 & 1.7 & 4.0 \\
        Ours {\scriptsize (Qwen2.5-VL-3B)} & 0.873 & 0.695 & 0.777 & 12.8 &18.1 & 6.3 & 36.02 & 1.5 & 1.1 & 1.6 & 2.6\\
        \bottomrule
    \end{tabular}}
    \end{center}
    \vspace{-.5cm}
    \caption{MLLM ablation with average performance and cost on GenAI-Bench [16]. Cost refers to GPU expenses unless otherwise noted.}
    \label{tab:llm_backbone_and_Cost_report}
\end{table*}
\addtocounter{table}{-1} 
\renewcommand{\thetable}{\arabic{table}} 

\vspace{0.2cm}

\begin{figure*}[h]
  \centering
  \includegraphics[width=1.0\linewidth]{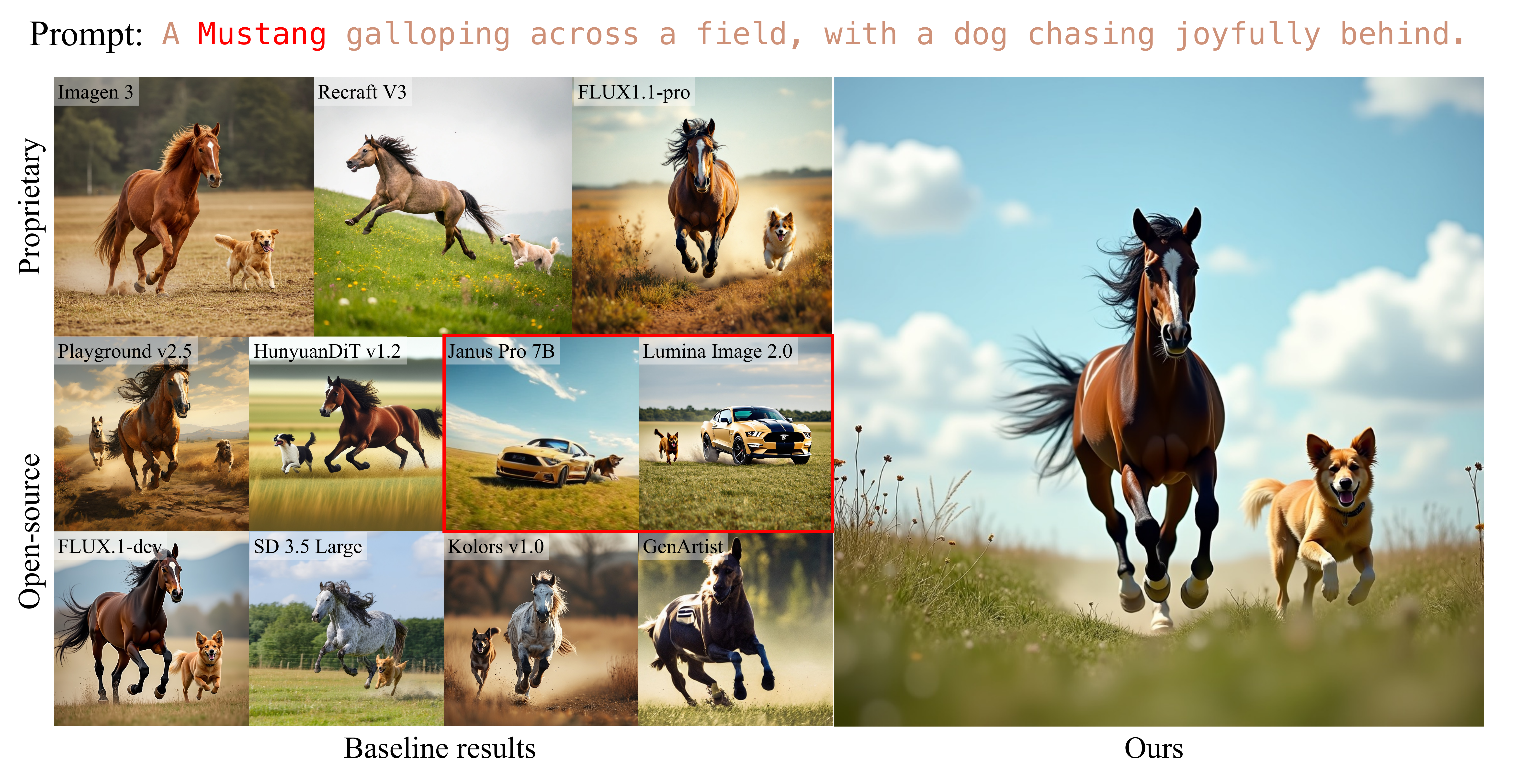}
  \captionsetup{skip=3pt}
   \caption{\textbf{Ambiguities sample.} 
   The prompt ``A Mustang galloping across a field, with a dog chasing joyfully behind'' is ambiguous; “Mustang” could mean a car or a horse. While Janus Pro 7B~\cite{janusPro7B_25} and Lumina Image 2.0~\cite{luminaImage_25} depict a Ford Mustang, others show a horse. Our Input Interpreter Agent resolves this by recognizing that “galloping” applies to horses, ensuring correct subject interpretation.}  
   \label{fig:ambiguity-mustang}
\end{figure*}

\section{Ablation study of MLLM backbones}
\label{sec:MLLM-ablation}
We evaluate open-sourced Mistral Small 3.1 24B~\cite{Mistral_small}, Qwen2.5-VL 7B and 3B~\cite{QWen2.5_VL} on L40S GPUs, with the 24B model using two GPUs and the others using one. 
Table~\ref{tab:llm_backbone_and_Cost_report} shows that performance is similar across model sizes: 
compared to GPT-4o-mini, 
7B model scores $-$0.2\% VQAScore at $-$40\% cost.

\section{Cost comparison}
\label{sec:cost_comparison}
In Table~\ref{tab:llm_backbone_and_Cost_report}, 
beyond the direct spending per image: \$0.005 for our method (using GPT-4o-mini) vs. \$0.04 for FLUX1.1-pro~\cite{flux24}, we also account for self-hosted hardware costs. Specifically, using an L40S GPU priced at \$11,250 and depreciated over five years with 144 hours of weekly usage results in a rate of \$0.30 per GPU-hour. Generating an average of 1,600 images, our method's $A_{gen}$ takes 19.72s/image, translating to \$0.0016/image. This amounts to only 16.59\% of FLUX1.1-pro’s cost, while achieving a +6\% improvement in VQAScore~\cite{VQAScore_ECCV24}. Compared to GenArtist~\cite{Genartist_NeurIPS24}, our method incurs 1.41x higher cost but delivers a +38\% gain in VQAScore~\cite{VQAScore_ECCV24}.

\section{More qualitative results}
\label{sec:qualitative-results}
\subsection{Ambiguous term}
Ambiguous terms in text-to-image prompts can lead to unintended or inconsistent image generation. When a term has multiple possible interpretations, different models may generate vastly different images, reflecting the ambiguity inherent in natural language. Take Fig.~\ref{fig:ambiguity-mustang} as an example.

\begin{figure*}[h]
  \centering
  \captionsetup{skip=3pt}
  \includegraphics[width=1.0\linewidth]{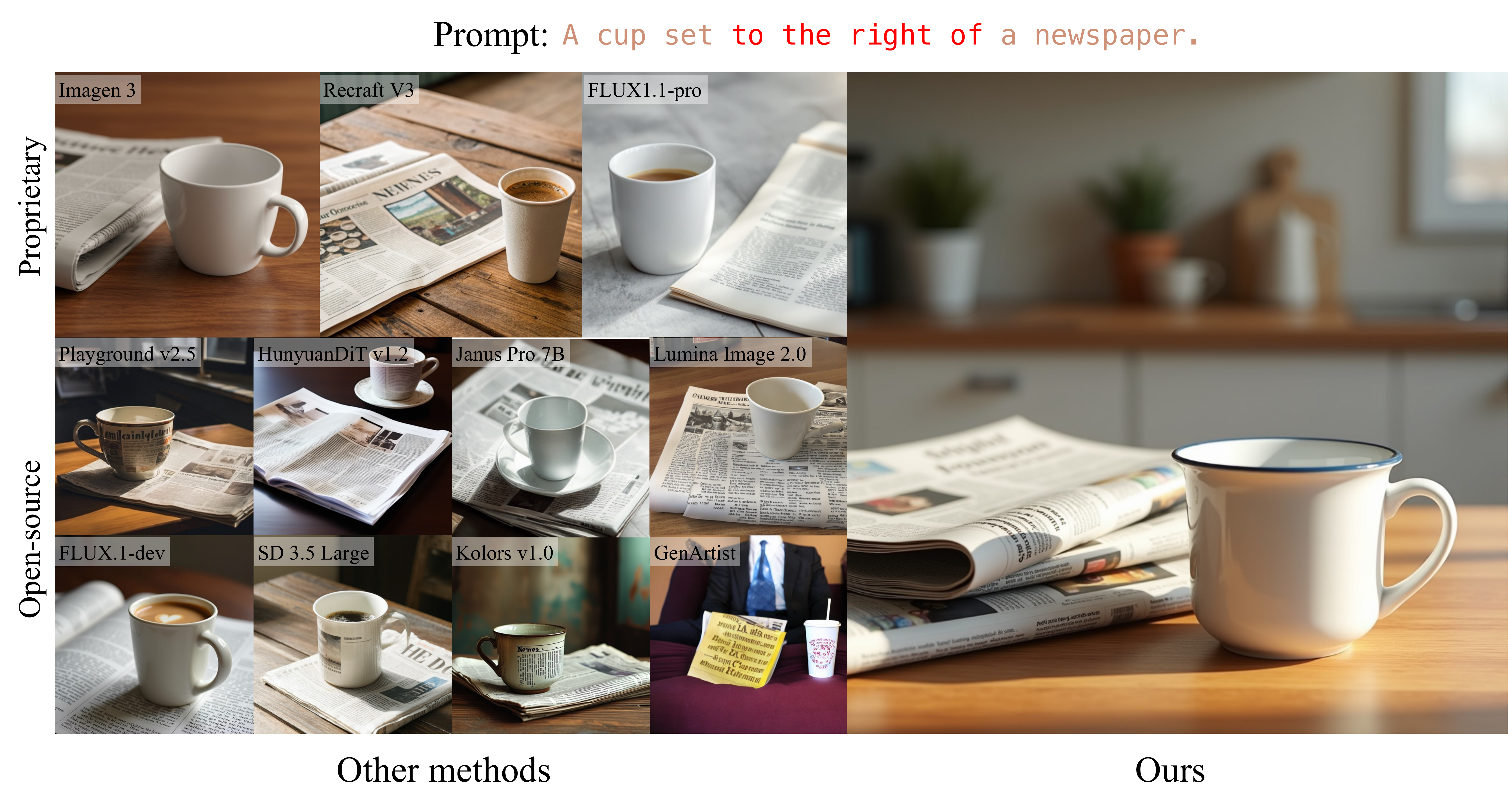}
   \caption{\textbf{Qualitative result in single-turn}: Demonstrate generation performance on positional relationship of two objects.}  
   \label{fig:cup-and-newspaper}
\end{figure*}
\begin{figure*}[h]
  \centering
  \captionsetup{skip=3pt}
  \includegraphics[width=1.0\linewidth]{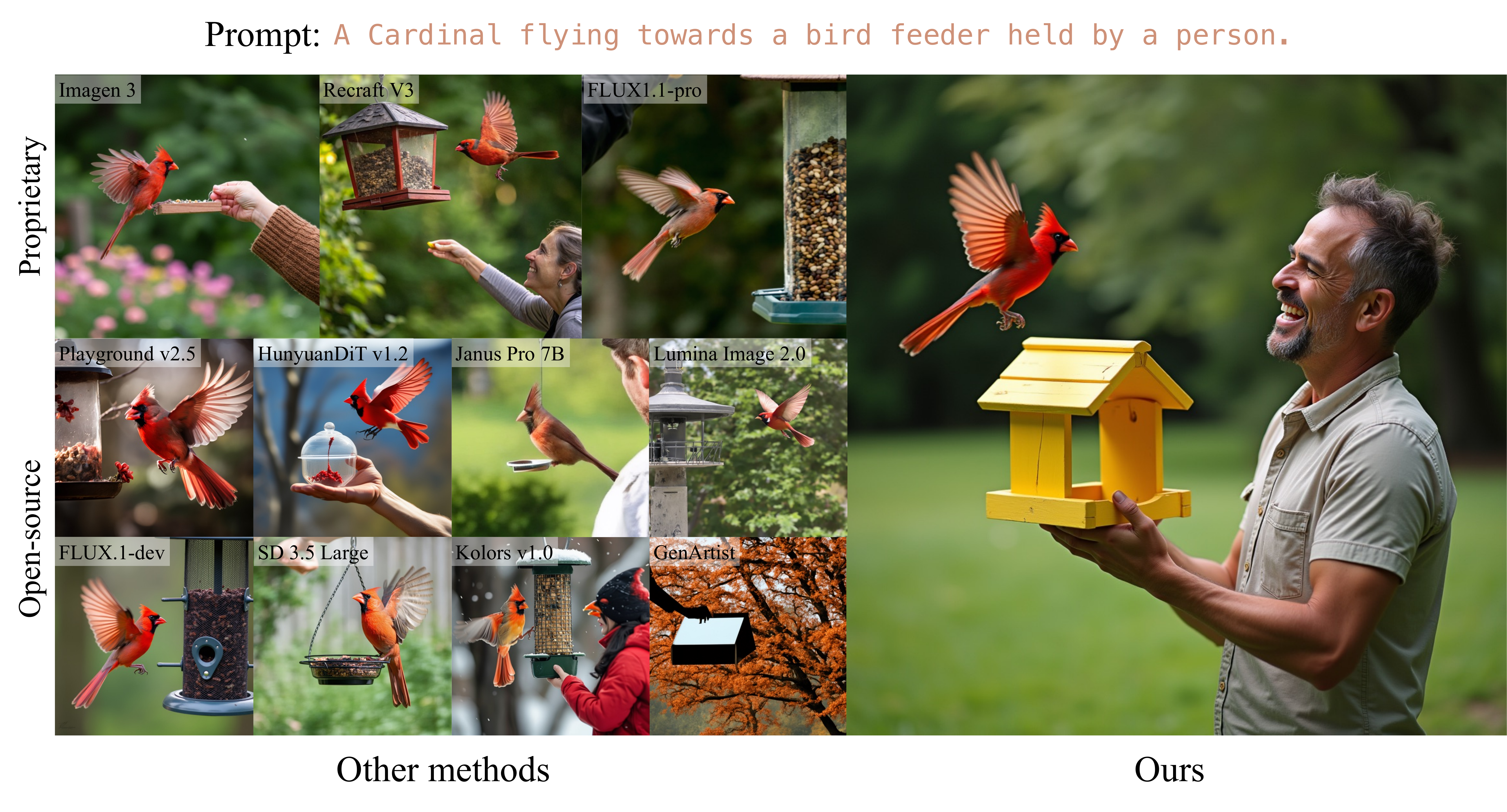}
   \caption{\textbf{Qualitative result in single-turn}: Demonstrate generation performance on action relationship of a bird and a human with given object (bird feeder).}
   \label{fig:bird-fly-to-human}
\end{figure*}
\begin{figure*}[h]
  \centering
  \captionsetup{skip=3pt}
  \includegraphics[width=1.0\linewidth]{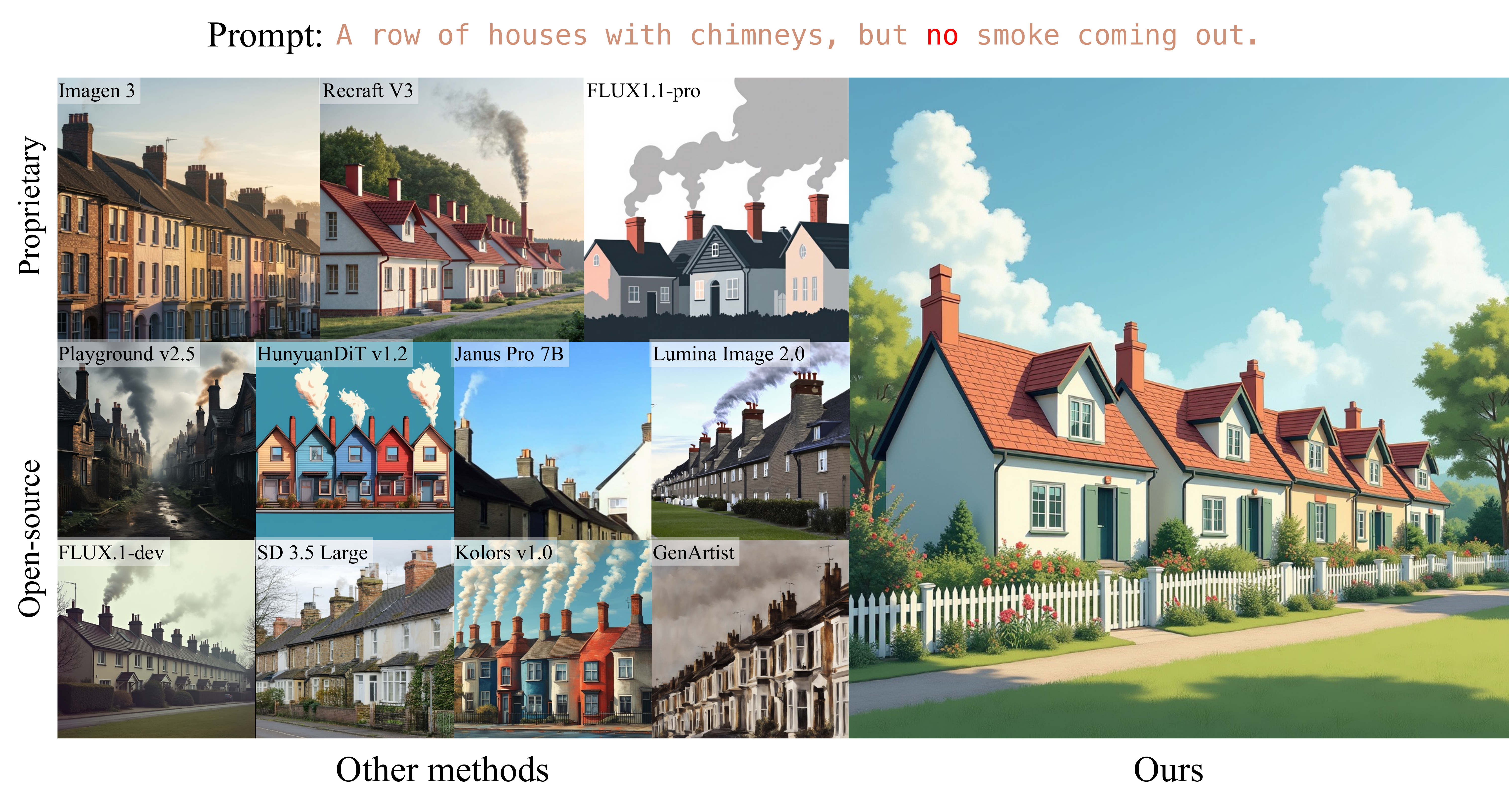}
   \caption{\textbf{Qualitative result in single-turn}: Demonstrate generation performance on logical negation of excluding smoke in the image.} 
   \label{fig:house-no-smoke}
\end{figure*}
\begin{figure*}[h]
  \centering
  \captionsetup{skip=3pt}
  \includegraphics[width=1.0\linewidth]{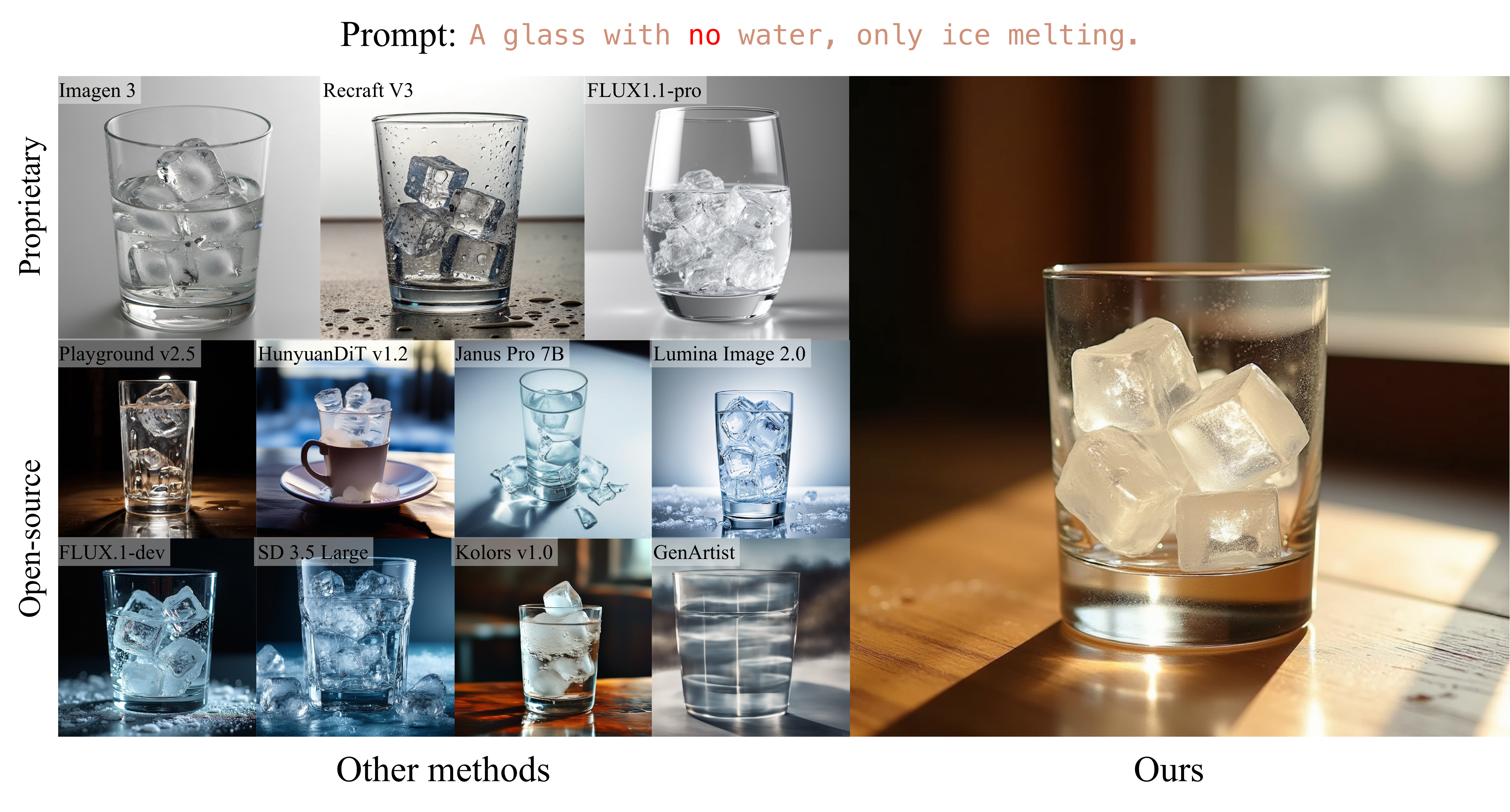}
   \caption{\textbf{Qualitative result in single-turn}: Demonstrate generation performance on logical negation of excluding water in the image.}   
   \label{fig:only-ice}
\end{figure*}
\begin{figure*}[h]
  \centering
  \captionsetup{skip=3pt}
  \includegraphics[width=1.0\linewidth]{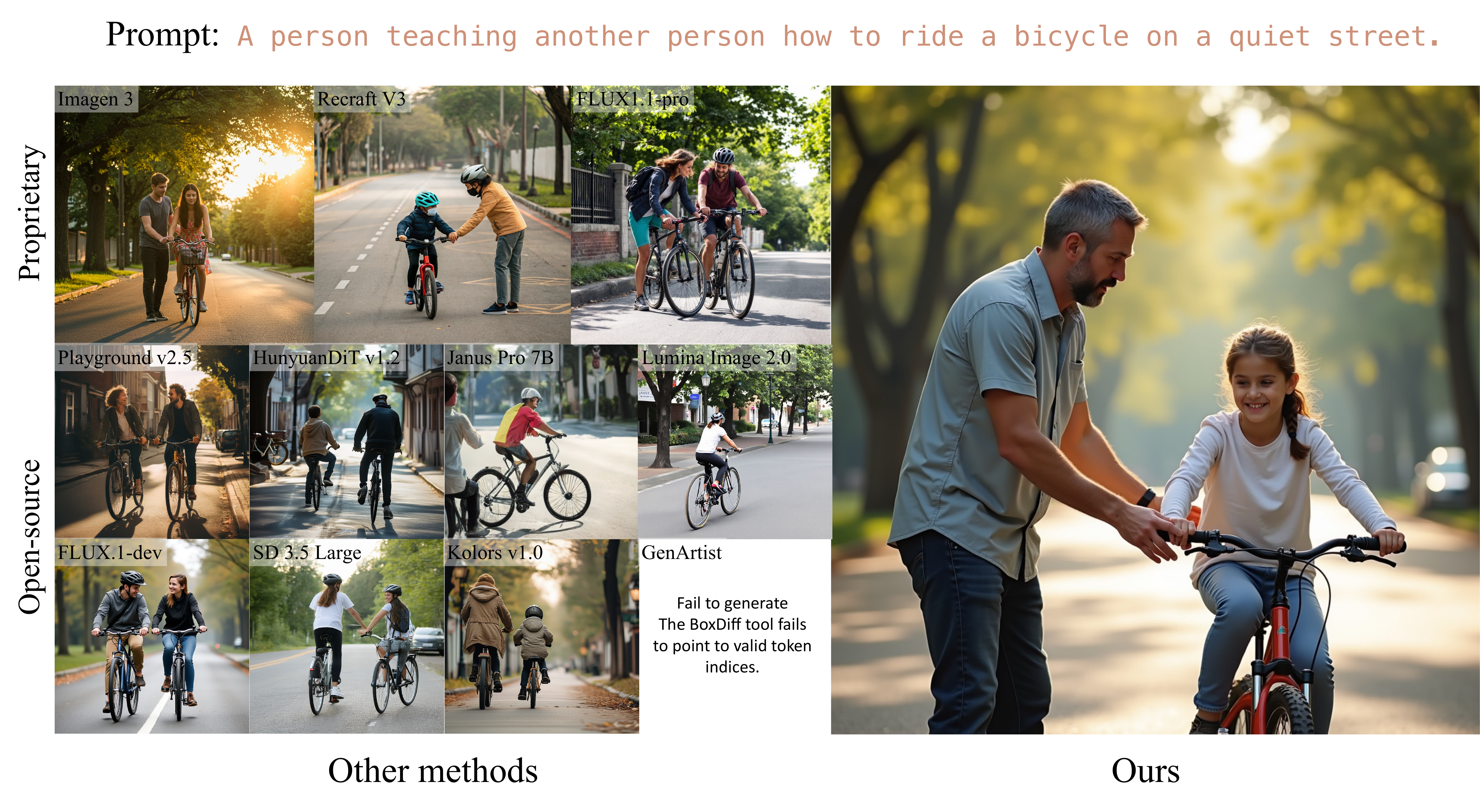}
   \caption{\textbf{Qualitative result in single-turn}: Demonstrate generation performance on action relationship of two persons in the image.}   
   \label{fig:teaching-bike}
\end{figure*}
\begin{figure*}[h]
  \centering
  \captionsetup{skip=3pt}
  \includegraphics[width=1.0\linewidth]{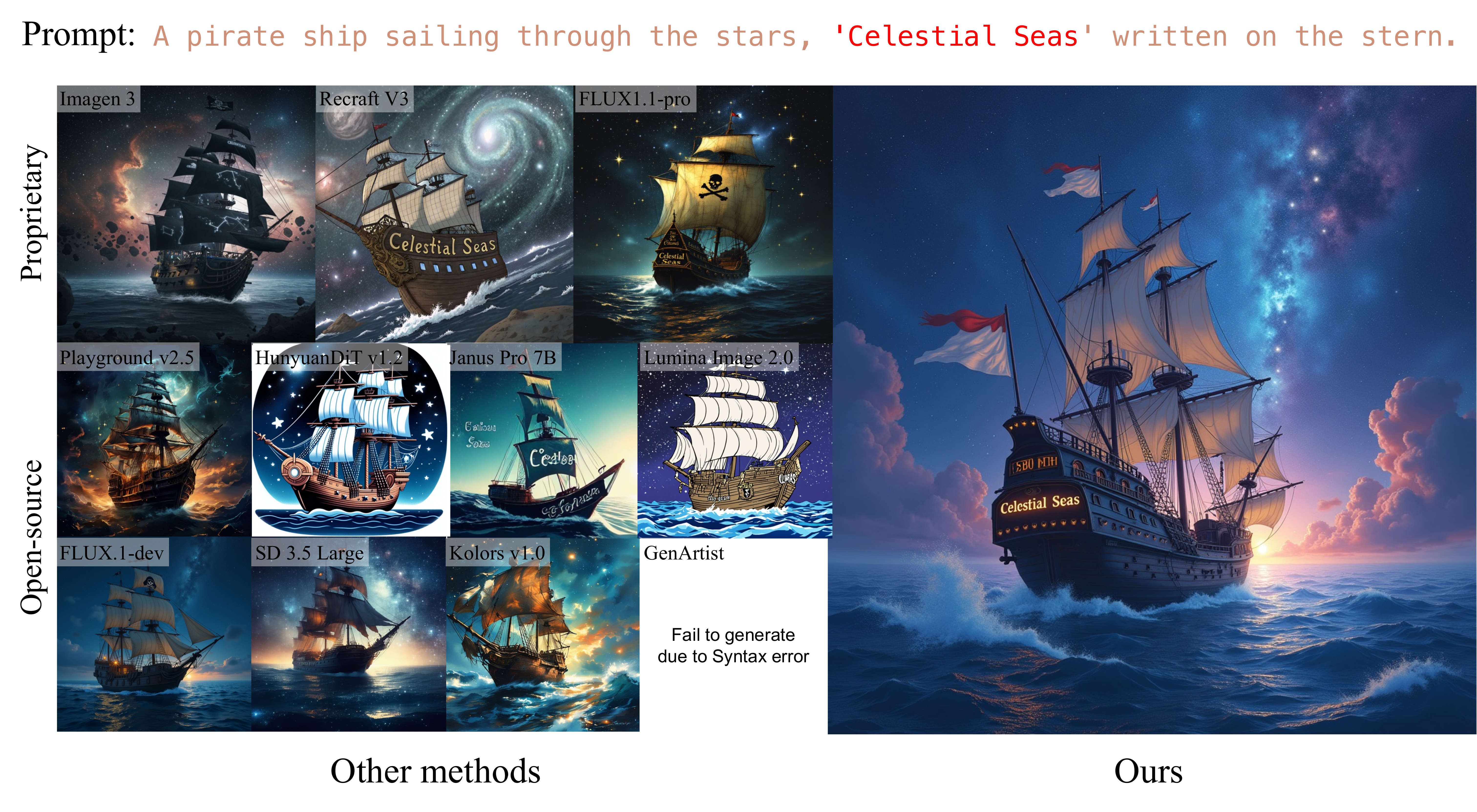}
   \caption{\textbf{Qualitative result in single-turn}: Demonstrate generation performance on including specific text in the specific region.}  
   \label{fig:private-ship}
\end{figure*}

\subsection{More results in single-turn: Figs.~\ref{fig:cup-and-newspaper},~\ref{fig:bird-fly-to-human},~\ref{fig:house-no-smoke},~\ref{fig:only-ice},~\ref{fig:teaching-bike},~\ref{fig:private-ship}.}

\subsection{More results in multi-turn}
\begin{figure*}[h]
  \centering
  \captionsetup{skip=3pt}
  \includegraphics[width=.7\linewidth]{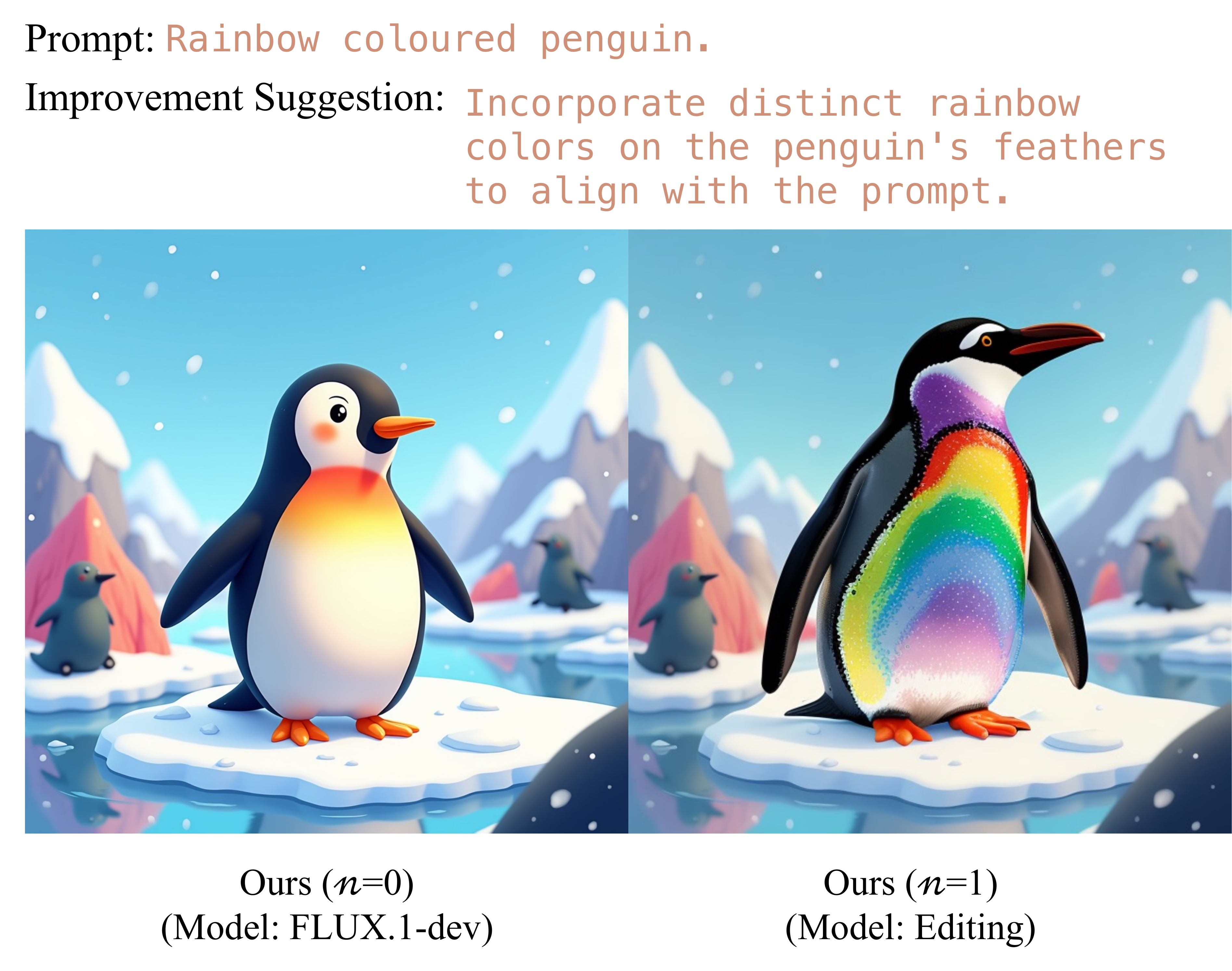}
   \caption{\textbf{Qualitative result in multi-turn}: Demonstrate enhancement performance on providing improvement suggestion and successfully modifying a specific region of the image automatically.} 
   \label{fig:auto-penguin}
\end{figure*}

\begin{figure*}[h]
  \centering
  \captionsetup{skip=3pt}
  \includegraphics[width=.7\linewidth]{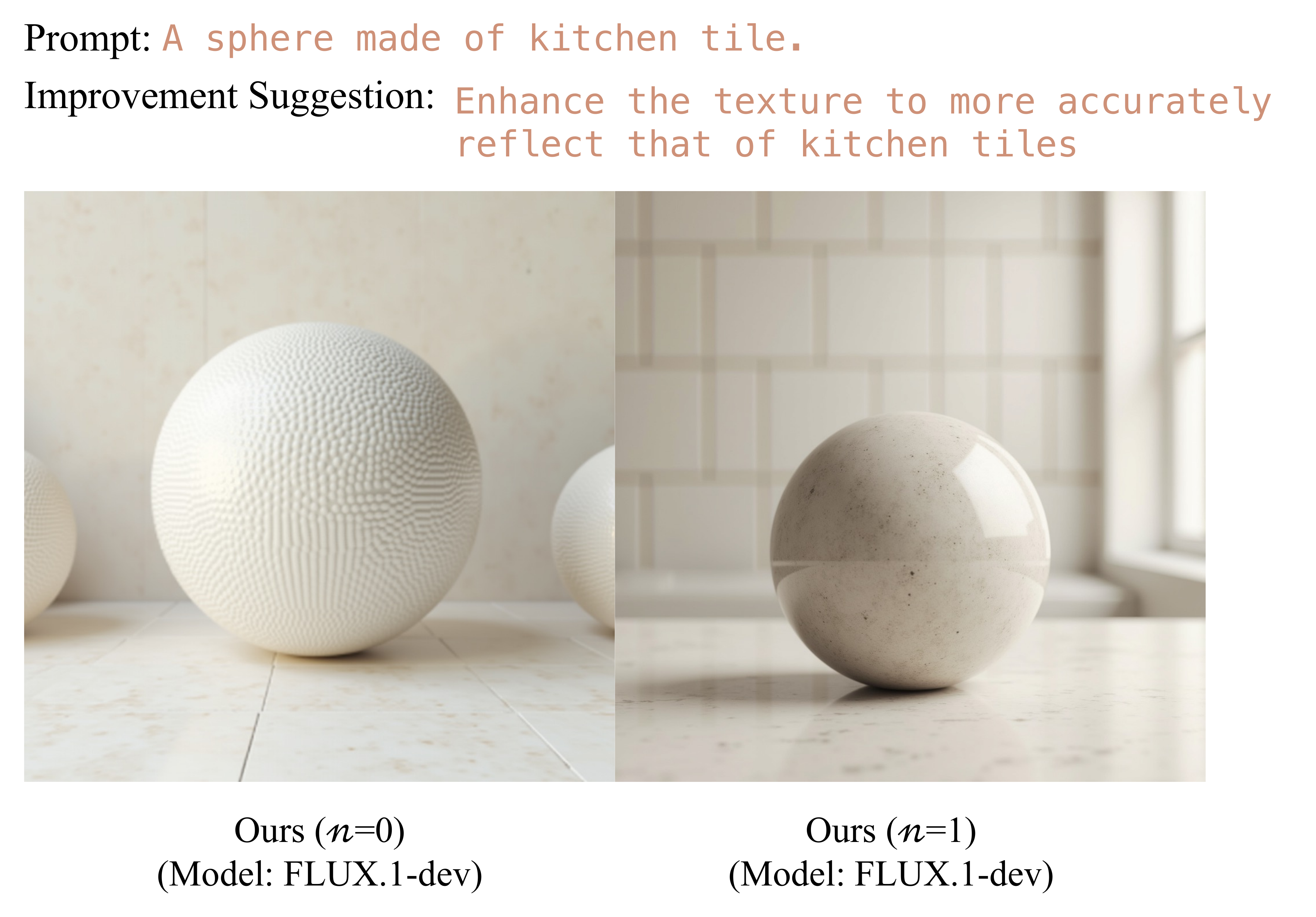}
   \caption{\textbf{Qualitative result in multi-turn}: Demonstrate enhancement performance on providing improvement suggestion and successfully modifying the texture of the image automatically.} 
   \label{fig:auto-kitchen-tile}
\end{figure*}

\subsubsection{Automatic: Figs.~\ref{fig:auto-penguin} and ~\ref{fig:auto-kitchen-tile}.}


\begin{figure*}[h]
  \centering
  \captionsetup{skip=3pt}
  \includegraphics[width=.7\linewidth]{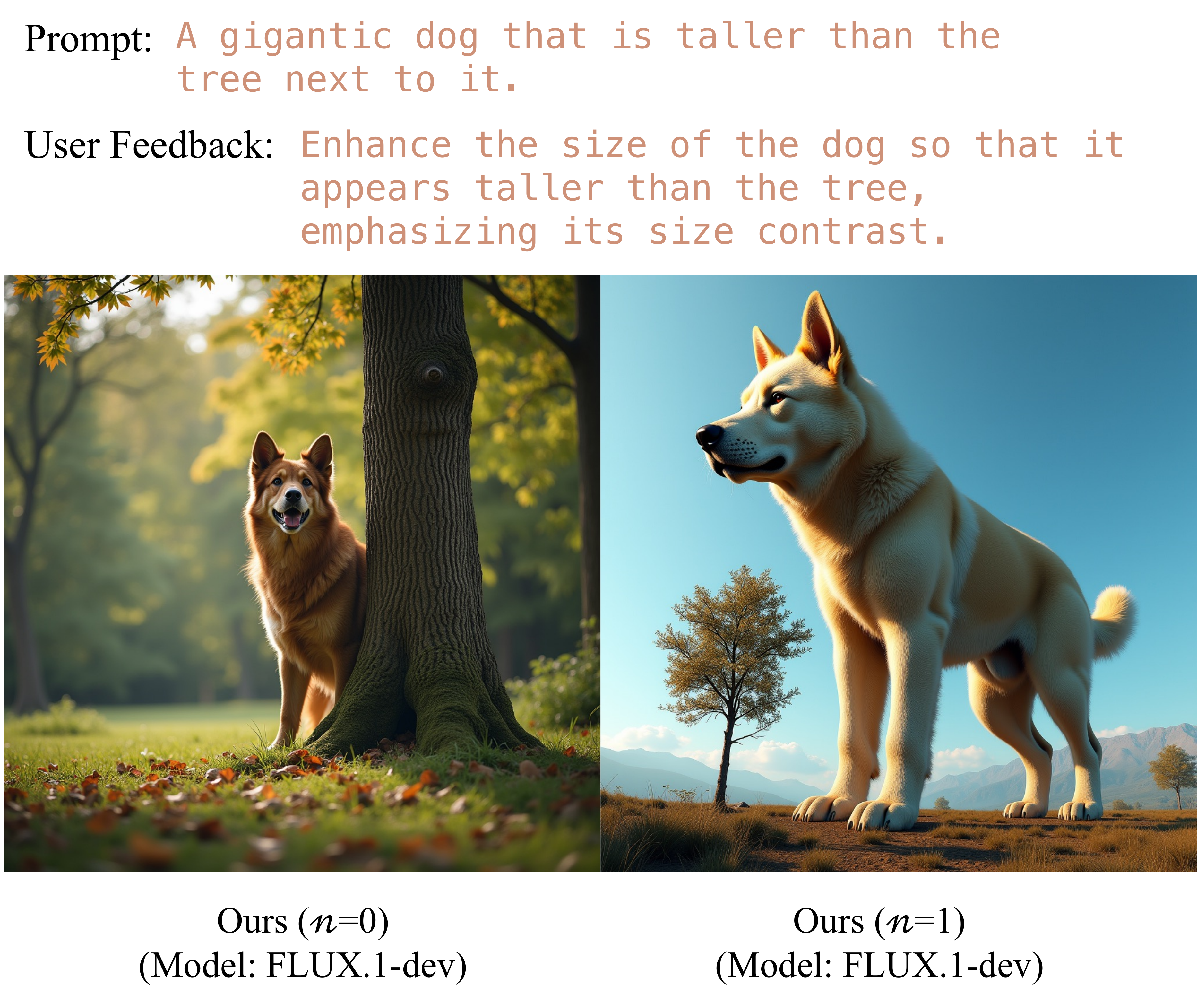}
   \caption{\textbf{Qualitative result in multi-turn}: Demonstrate enhancement performance on including user feedback and successfully modifying the size contrast of dog and tree.}  
   \label{fig:hilp-giant-dop}
\end{figure*}

\begin{figure*}[h]
  \centering
  \captionsetup{skip=3pt}
  \includegraphics[width=.7\linewidth]{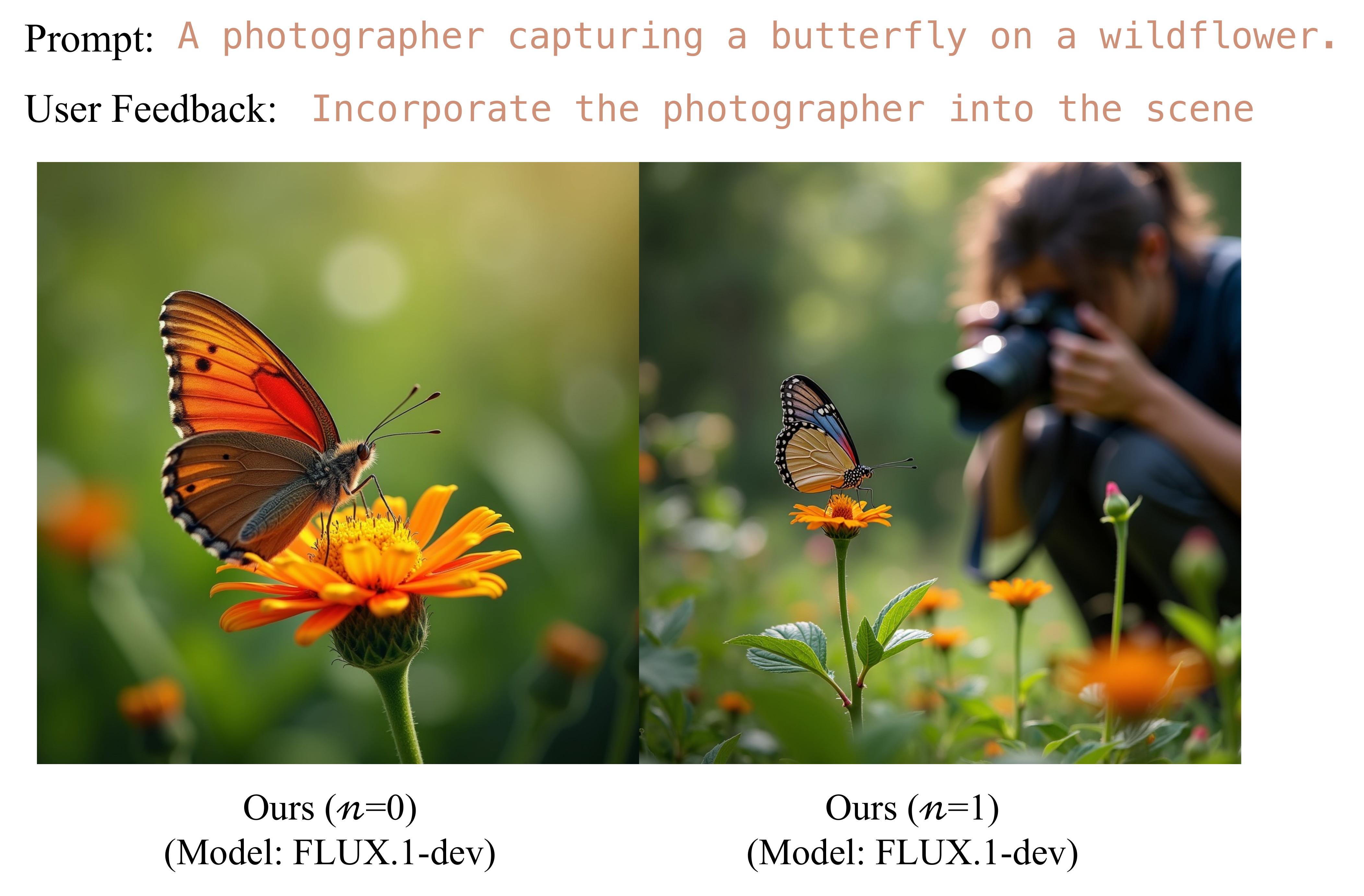}
   \caption{\textbf{Qualitative result in multi-turn}: Demonstrate enhancement performance on including user feedback and successfully including the photographer into the scene.} 
   \label{fig:hilp-photographer}
\end{figure*}

\subsubsection{Human-in-the-loop: Figs.~\ref{fig:hilp-giant-dop} and ~\ref{fig:hilp-photographer}.}

\begin{figure}
  \centering
  \includegraphics[width=\linewidth]{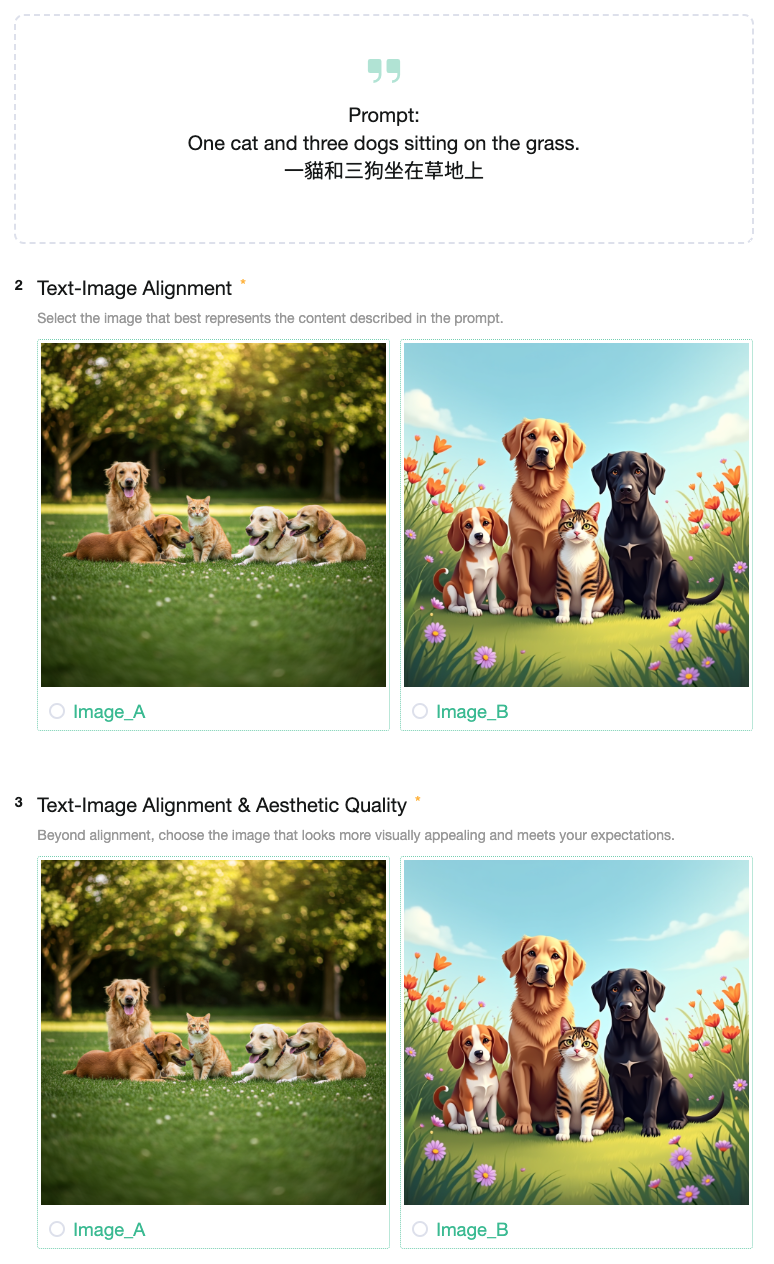}
   \caption{\textbf{The screenshot of user study website.}}  
   \label{fig:user-study-screenshot}
\end{figure}

\section{User study website screenshot: Fig.~\ref{fig:user-study-screenshot}.}

\end{document}